\UseRawInputEncoding
\documentclass[11pt]{article}
\usepackage[preprint]{acl}
\usepackage{fontawesome5}
\usepackage{times}
\usepackage{latexsym}
\usepackage[T1]{fontenc}
\usepackage[utf8]{inputenc}
\usepackage{microtype}
\usepackage{amsmath}
\usepackage{amssymb}
\usepackage{booktabs}
\usepackage{multirow}
\usepackage[table]{xcolor}
\usepackage{tabularx}
\usepackage{inconsolata}
\usepackage{graphicx}
\usepackage{subcaption}
\usepackage{amssymb}

\definecolor{linkblue}{HTML}{2A2D6E}

\newcommand{\keyreview}{%
  \raisebox{-0.15ex}{%
    \textcolor{red!80!black}{\Large$\bigstar$}%
  }\hspace{0.3em}%
}
\usepackage{listings}
\usepackage[most]{tcolorbox}

\usepackage{dashrule}
\usepackage{tcolorbox}

\tcbset{
  casebox/.style={
    width=\linewidth,
    boxrule=0.3pt,
    arc=1mm,
    boxsep=0pt,
    left=1.2mm,
    right=1.2mm,
    top=1mm,
    bottom=1mm,
    before skip=3pt,
    after skip=3pt
  }
}

\newtcolorbox{memoryinfo}{
  casebox,
  colback=green!6,
  colframe=black
}

\newtcolorbox{memorychat}{
  casebox,
  colback=blue!4,
  colframe=black
}

\newtcolorbox{memoryreview}{
  casebox,
  colback=red!10,
  colframe=black
}

\newcommand{\memorydivider}{%
  \par\smallskip
  \noindent\hdashrule{\linewidth}{0.4pt}{2pt 2pt}%
  \par\smallskip
}   

\newtcblisting[auto counter]{examplebox}[2][]{
  enhanced,
  breakable,
  listing only,
  title={Example~\thetcbcounter: #2},
  colback=gray!3,
  colframe=black!70,
  colbacktitle=blue!12,
  coltitle=black,
  fonttitle=\small\bfseries,
  boxrule=0.8pt,
  arc=1.5mm,
  left=2mm,
  right=2mm,
  top=1.5mm,
  bottom=1.5mm,
  before skip=8pt,
  after skip=8pt,
  listing options={
    basicstyle=\footnotesize\ttfamily,
    columns=fullflexible,
    keepspaces=true,
    breaklines=true,
    breakatwhitespace=false,
    showstringspaces=false,
    numbers=none,
    tabsize=2
  },
  #1
}

\usepackage{listings}
\tcbuselibrary{listings,breakable}

\newtcblisting[auto counter]{promptbox}[2][]{
  enhanced,
  breakable,
  listing only,
  title={Prompt~\thetcbcounter: #2},
  fonttitle=\small\bfseries,
  coltitle=black,
  colback=gray!3,
  colbacktitle=blue!12,
  colframe=black,
  boxrule=0.5pt,
  arc=1mm,
  left=1.5mm,
  right=1.5mm,
  top=1.5mm,
  bottom=1.5mm,
  before skip=8pt,
  after skip=8pt,
  listing options={
    basicstyle=\footnotesize\ttfamily,
    columns=fullflexible,
    keepspaces=true,
    breaklines=true,
    breakatwhitespace=false,
    showstringspaces=false,
    numbers=none,
    tabsize=2
  },
  #1
}

\newtcblisting[auto counter]{memoryexample}[2][]{
  enhanced,
  breakable,
  listing only,
  listing engine=listings,
  listing options={
    language={},
    basicstyle=\ttfamily\footnotesize,
    breaklines=true,
    breakatwhitespace=false,
    columns=fullflexible,
    keepspaces=true,
    showstringspaces=false,
    mathescape=false,
    texcl=false,
    upquote=true
  },
  title={Memory~\thetcbcounter: #2},
  fonttitle=\bfseries,
  coltitle=black,
  colbacktitle=blue!8,
  colback=gray!3,
  colframe=black,
  boxrule=0.5pt,
  arc=1mm,
  left=1.5mm,
  right=1.5mm,
  top=1mm,
  bottom=1mm,
  #1
}

\newcommand{\method}{\textsc{MemComposer}}

\newcommand{\huggingface}{\raisebox{-1.5pt}{\includegraphics[height=1.05em]{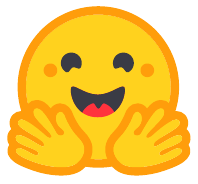}}}
\newcommand{\github}{\raisebox{-1.5pt}{\includegraphics[height=1.05em]{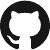}}}

\newcommand{\qwenicon}{\includegraphics[height=1.15em]{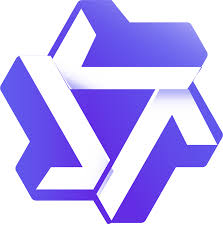}}
\newcommand{\mistralicon}{\includegraphics[height=1.15em]{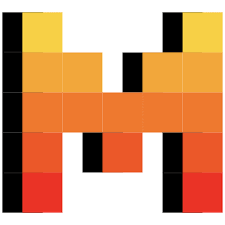}}
\newcommand{\ci}[2]{{\scriptsize[#1,\,#2]}}

\title{\method{}: Learning to Assemble Memories for LLM Agents via Reinforcement Learning}

\author{
 \textbf{Yixuan Xie\textsuperscript{1}}\hspace{0.7em}
 \textbf{Xueqiang Xu\textsuperscript{1,\dag}}\hspace{0.7em}
 \textbf{Xinyi Fan\textsuperscript{1}}\hspace{0.7em}
 \textbf{Xinxi Lyu\textsuperscript{1}}\hspace{0.7em}
 \textbf{Tao Feng\textsuperscript{1}}\hspace{0.7em}
 \textbf{Xing Fan\textsuperscript{2}} \\
 \textbf{Chenlei Guo\textsuperscript{2}}\hspace{0.7em}
 \textbf{Jie Hao\textsuperscript{2}} \hspace{0.7em}
 \textbf{Shaowen Wang\textsuperscript{1}}\hspace{0.7em}
 \textbf{Jiashuo Sun\textsuperscript{1,\dag}}\hspace{0.7em}
 \textbf{Jiawei Han\textsuperscript{1,\dag}} \\[2pt]
 \textsuperscript{1}University of Illinois Urbana-Champaign \quad
 \textsuperscript{2}Amazon \\
 \texttt{\{xie39,hanj\}@illinois.edu} \quad
 \textsuperscript{\dag}Corresponding authors \\[4pt]
 \github\ \href{https://github.com/Makr-Xie/MemComposer}{\texttt{Code}}
 \hspace{0.7em}
 \huggingface\ \href{https://huggingface.co/collections/Makrrr/memcomposer}{\texttt{Model \& Data}}
}

\begin{document}
\maketitle

\begin{abstract}
Agent memory enables large language model (LLM) agents to reuse past experiences without updating their parameters. However, semantic similarity alone does not guarantee memory usefulness: downstream performance depends on how memories are selected and presented to the agent. We propose \textbf{\method{}}, a memory composer that jointly examines the current query and retrieved candidates to decide which experiences to include, how many to select, and how to order them. We train the composer using Experience Policy Optimization (ExPO), a GRPO-based framework that evaluates assembled memory contexts through complete agent interactions and uses task-success feedback to optimize the assembly policy. Across three $\tau^2$-bench domains with two frozen executors, \method{} outperforms the strongest baseline for each executor while reducing average memory-token usage by 16.9\% and 4.9\%, respectively. The learned policy also transfers to unseen domains, executors, and memory banks, and benefits from memories generated by stronger models without further training. These results demonstrate the value of learning query-specific memory assembly using downstream task feedback.
\end{abstract}

\section{Introduction}
\begin{figure}[t]
    \centering
    \includegraphics[width=\linewidth]{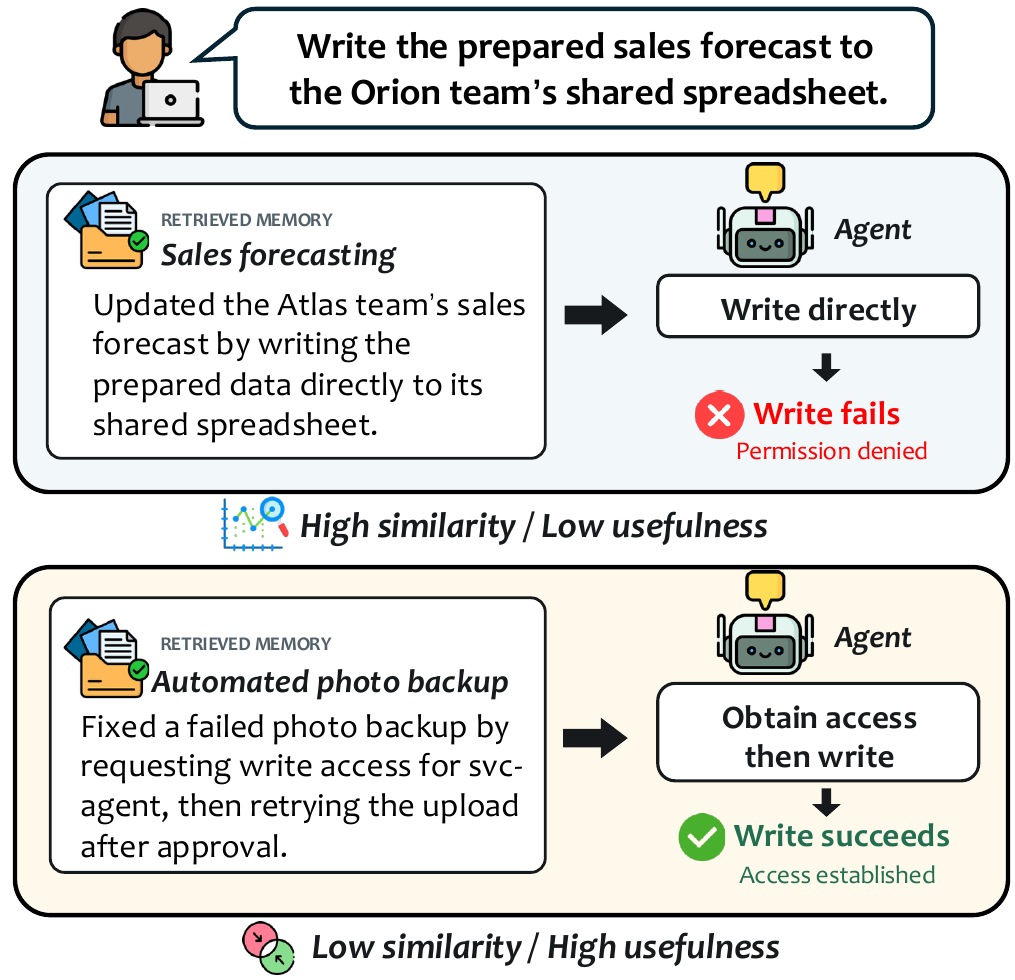}
    \caption{\textbf{Semantic Similarity $\neq$ Memory Usefulness.} The more similar memory misses the permission issue; the less similar one leads to success.}
    \label{fig:similarity_and_usefulness}
\end{figure}

\begin{figure*}[t]
  \centering
  \begin{subfigure}[b]{0.60\textwidth}
    \centering
    \includegraphics[width=\linewidth]{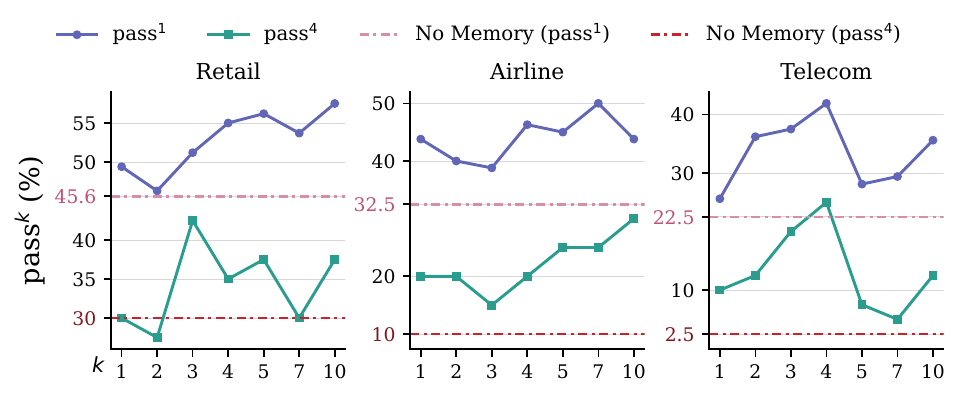}
    \caption{Effect of memory count}
    \label{fig:motivation_1}
  \end{subfigure}\hfill
  \begin{subfigure}[b]{0.38\textwidth}
    \centering
    \includegraphics[width=\linewidth]{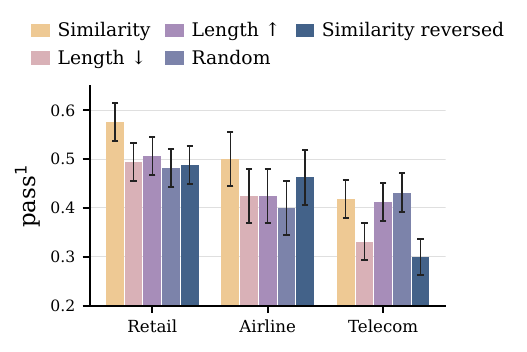}
    \caption{Effect of ordering}
    \label{fig:motivation_2}
  \end{subfigure}

  \caption{
    \textbf{Both the number and the order of memory matter.}
    Frozen Qwen3-4B-Instruct agent on $\tau^2$-bench
    pass$^1$ averages $c/n$ across tasks; pass$^4$ averages $\binom{c}{4}/\binom{n}{4}$, where $c$ of $n$ trials succeed. Dashed lines denote no-memory performance.
    (a) Increasing $k$ does not consistently improve performance, and the best $k$ differs across domains.
    (b) Reordering the same memory set changes per-task outcomes, though no single order is best on average.}
  \label{fig:motivation}
\end{figure*}

\begin{figure*}[t]
  \centering
  \includegraphics[width=\textwidth]{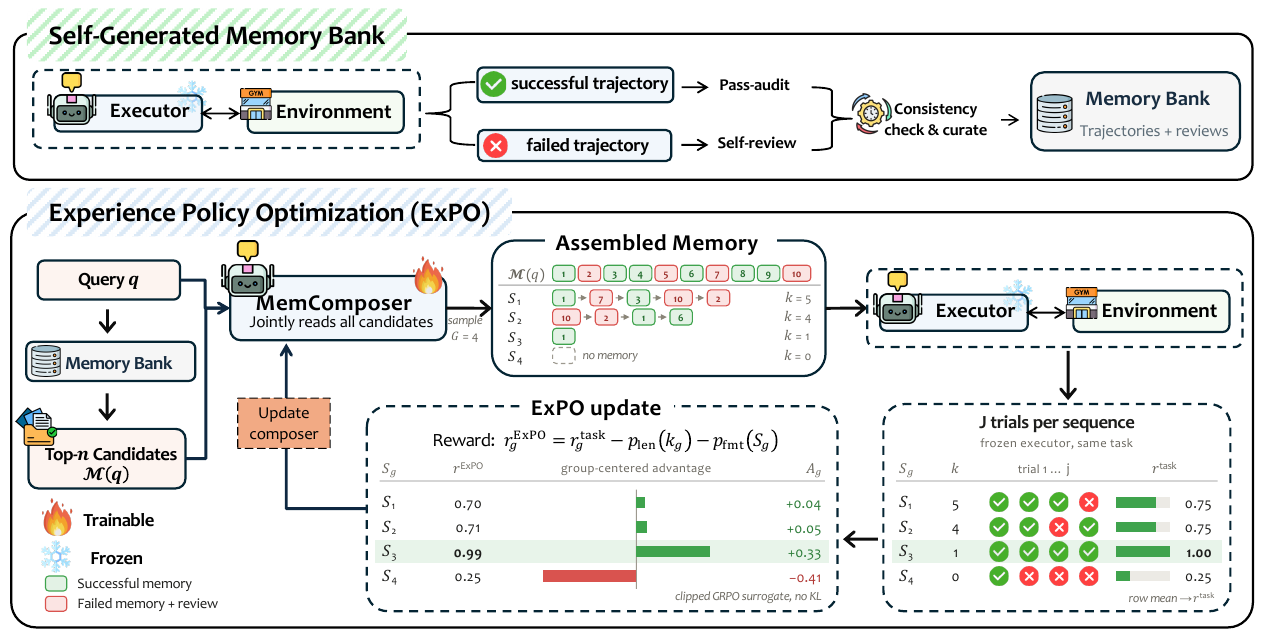}
  \caption{Overview of \method{}. 
  \textbf{Top:} Self-generated trajectories are reviewed and curated into a fixed memory bank.
  \textbf{Bottom:} The composer samples $G$ variable-size, ordered memory
  sequences per query, each evaluated over $J$ executor trials; ExPO updates
  the composer with their group-relative advantages, while the executor and
  memory bank stay fixed.
  }
  \label{fig:overview}
\end{figure*}

Agent memory augments large language models (LLMs) with reusable experience~\citep{zhang2024survey, hu2025memory}, skills~\citep{wang2023voyager, zheng2024synapse, ouyang2026skillos}, and knowledge~\citep{park2023generative, zhong2024memorybank, packer2023memgpt, xu2025amem, chhikara2025mem0} without updating model parameters, and has shown promising potential for agent continuous learning~\citep{shinn2023reflexion,  suzgun2026dynamic, ouyang2026reasoningbank, zhang2026ace}. A typical agent memory system stores past experiences in an external bank,
retrieves candidate memories for a new task, and inserts them into the
agent's context to guide its subsequent decisions. However, the effectiveness of an agent's memory system depends on how useful the retrieved memories are for the current task. 
Recent work on agentic memory~\citep{wang2025awm, zhao2024expel} has primarily focused on improving memory representations, while still relying on similarity-based methods, such as sparse retrieval~\citep{robertson2009bm25} and embedding-based retrieval~\citep{karpukhin2020dpr,izacard2022contriever, zhang2025qwen3embedding}, for retrieved memories.

Although such methods are widely used in knowledge-intensive scenarios such as retrieval-augmented generation (RAG)~\citep{lewis2020rag, guu2020realm, gao2023ragsurvey} and offer a natural starting point, similarity alone does not guarantee that a retrieved memory is useful. As illustrated in Figure~\ref{fig:similarity_and_usefulness}, a memory describing semantically less similar content provides the crucial information while the similar memory does not. 
Recent work has accordingly used downstream execution feedback to learn memory utility or optimize memory-use policies~\citep{zhang2026memrl, zhou2025memento, yan2026memoryr1}. 
Closing this gap therefore requires learning how candidate memories jointly affect downstream task performance, enabling the model to assemble a variable-length memory sequence tailored to the current query.

To better understand how memory assembly affects downstream task performance, we examine two key decisions: how many memories to include and in what order to present them. Figure~\ref{fig:motivation_1} shows that performance varies non-monotonically with memory count, with different domains favoring different retrieval budgets. Figure~\ref{fig:motivation_2} further shows that even when the selected memories remain unchanged, their presentation order affects task outcomes, consistent with order sensitivity observed in in-context learning~\citep{lu2022fantastically}. 
Together, these observations suggest that constructing a useful memory context requires decisions beyond ranking memories by similarity. To address this problem, we introduce \textbf{\method{}}, a memory composer that jointly examines the task query and retrieved candidates to select an ordered subset of variable size. 
We train the composer using Experience Policy Optimization (ExPO), a GRPO-based framework that evaluates assembled contexts through complete agent interactions.

By learning to assemble useful memory combinations, \method{} outperforms the strongest baseline for each agent while using fewer memory tokens per task: 26.9\% fewer for Qwen3-4B-Instruct and 35.2\% fewer for Ministral-3-8B-Instruct on \textsc{retail}, and 16.9\% and 4.9\% fewer, respectively, across all domains. Furthermore, the learned composer demonstrates strong transferability to unseen domains, executors, and memory banks. 
Notably, \method{} continues to improve performance when supplied
with memories generated by stronger models, without further training
(Figure~\ref{fig:memory}). This suggests that it learns a transferable
ability to assess which memories are useful for downstream tasks.

Our contributions are summarized as follows:
\begin{itemize}
    \item We propose \method{}, a memory composer that jointly selects and orders a variable-size subset of retrieved experiences for a frozen LLM agent.

    \item We develop Experience Policy Optimization (ExPO), a GRPO-based training framework that evaluates assembled contexts through complete agent interactions and combines empirical task success with selection-size and format penalties.
    
    \item Experiments on $\tau^2$-bench demonstrate improved task success and memory-token efficiency with two frozen executors, together with strong transferability across domains, executors, and memory sources.
\end{itemize}

\section{Related Work}
\label{sec:related-works}

\subsection{Memory for LLM Agents}
Agent memory lets LLM agents reuse past experience as trajectories and
insights \citep{zhao2024expel, ouyang2026reasoningbank}, workflows and
skills \citep{wang2025awm, wang2023voyager, ouyang2026skillos,
zhang2026usefulmemory}, or general-purpose stores
\citep{packer2023memgpt, chhikara2025mem0, xu2025amem}, but memories are
typically recalled by embedding similarity. Methods that learn memory
recall or management from task outcomes \citep{zhang2026memrl,
zhou2025memento, yan2026memoryr1} score or operate on memories
individually, whereas \method{} decides the assembled context as a
whole: which memories, how many, and in what order.

\subsection{Learning Context Selection from Downstream Feedback}
Demonstration selection for in-context learning \citep{rubin2022epr, li2023udr, wang2024llmr, ye2023ceil, liu2024se2, zhang2022active, scarlatos2023reticl} and passage reranking for retrieval-augmented generation \citep{zhuang2025rankr1, zhang2025rearank, jiang2025s3, sun2026barrag, sun2025dynamicrag} have been learned from downstream feedback, in some cases jointly choosing the subset and its order.
These methods rely on cheap per-query signals such as gold-output likelihood, relevance labels, or the quality of a single answer. In our setting, candidates are agent trajectories with hindsight reviews, and a context is scored only after a multi-turn interaction with tools and a simulated user, yielding a sparse, stochastic reward that ExPO estimates over repeated executor trials. 

Due to space limits, we provide an extended discussion in Appendix~\ref{app:related-works}.

\section{Method}
In this section, we first formulate memory assembly (\S\ref{sec:formulation}), then describe how the experience memory is built (\S\ref{sec:memory}) and how \method{} assembles it (\S\ref{sec:method}). Finally, we introduce Experience Policy Optimization (ExPO), which trains \method{} with task-outcome rewards (\S\ref{sec:training}). Figure~\ref{fig:overview} gives an overview of our framework.

\subsection{Problem Formulation}
\label{sec:formulation}
Let $\pi_\theta$ denote the composer policy and $\pi_e$ the executor policy.
Given a task query $q$, let $\mathcal{M}(q) = \{m_1, \dots, m_n\}$ be the top-$n$ candidate memories returned by a similarity-based retriever from a memory bank $\mathcal{B}$ (\S\ref{sec:memory}).
The composer $\pi_\theta$ samples an ordered subset $S = (m_{i_1}, \dots, m_{i_k})$ of distinct candidates, with $0 \le k \le n$, jointly deciding which memories to include and their order.
Conditioned on $(q, S)$, the executor completes the task through multi-turn interaction with the environment, which returns a binary success reward $r \in \{0,1\}$.

Let $R(q, S)$ denote the executor's expected success reward when
using the memory sequence $S$ for query $q$.
We train the composer to maximize
\begin{equation}
  \max_{\theta}\;
  \mathbb{E}_{q \sim \mathcal{D}_{\mathrm{train}}}
  \mathbb{E}_{S \sim \pi_\theta(\cdot \mid q,\mathcal{M}(q))}
  \big[R(q,S)\big],
  \label{eq:objective}
\end{equation}

\subsection{Self-Generated Memory Bank}
\label{sec:memory}
For each domain and agent backbone, we collect $N$ independent trajectories $\tau$ per training task $q \in \mathcal{D}_{\mathrm{train}}$ using the frozen executor $\pi_e$ and record the outcome $r = r(q,\tau)$.
The same model then produces a hindsight review $v$ for each trajectory--outcome pair $(\tau,r)$, without modifying the original outcome $r$.

We validate the reviews with deterministic consistency checks and curate the validated memory $m=(\tau,r,v)$ into a memory bank $\mathcal{B}$, shared by all methods evaluated with the same domain and agent backbone.

Appendix~\ref{app:memory} provides the review prompts and scripts,
validation and curation rules, parameter settings, and final
memory bank sizes.

\subsection{\method{}}
\label{sec:method}
Given a task query $q$, we retrieve the top-$n$ candidate memories $\mathcal{M}(q)$ from memory bank $\mathcal{B}$ using a fixed similarity-based retriever.

The composer $\pi_\theta$ reads $q$ and all candidate memories jointly to generate a memory sequence:
\begin{equation}
  S = (m_{i_1}, \dots, m_{i_k})
  \sim \pi_\theta\big(\cdot \mid q, \mathcal{M}(q)\big).
\end{equation}
For each task sample, both the length $k$ and the ordering of $S$
are dynamically determined by the composer, and are learned from the task outcomes using Experience Policy Optimization (\S\ref{sec:training}).
The implementation details for \method{} are provided in Appendix~\ref{app:method}.

\subsection{Experience Policy Optimization}
\label{sec:training}
We introduce Experience Policy Optimization (ExPO) to optimize Eq.~\ref{eq:objective}, training the composer from the downstream utility of the memory contexts it assembles.
ExPO builds on Group Relative Policy Optimization (GRPO)~\citep{shao2024deepseekmath}, but evaluates each memory sequence through complete downstream task interactions and combines its empirical task success rate with penalties on selection size and output format.
Because each sequence is executed with its memories in the order given, this reward shapes not only which memories are selected but also how they are ordered, without a separate ordering objective.

\paragraph{Reward Design.}
\label{sec:reward-function}
Following prior work~\citep{jiang2025s3, zhang2026memrl, sun2026barrag}, we use downstream task outcomes to assess the utility of assembled memory contexts.
Since agentic execution is stochastic~\citep{kapoor2024agents, yao2024tau}, a single rollout provides a noisy estimate of a memory sequence's utility. We therefore evaluate each sampled sequence $S_g$ over $J$ trials with the frozen executor and define its task-outcome reward as the empirical success rate:
\begin{equation}
  r_g^{\mathrm{task}}
  = \frac{1}{J}\sum_{j=1}^{J} r_{g,j},
  \label{eq:task-reward}
\end{equation}
where $r_{g,j} \in \{0,1\}$ is the outcome of trial $j$.

We additionally apply a selection-size penalty $p_{\mathrm{len}}(k_g)$, where $k_g = |S_g|$, to discourage unnecessarily large memory contexts.
The format penalty $p_{\mathrm{fmt}}(S_g)$ is applied when the composer violates the required reasoning and answer structure.
The resulting ExPO reward is
\begin{equation}
  r_g^{\mathrm{ExPO}}
  = r_g^{\mathrm{task}}
  - p_{\mathrm{len}}(k_g)
  - p_{\mathrm{fmt}}(S_g).
  \label{eq:expo-reward}
\end{equation}
This reward balances downstream task success, selection size, and output validity.
Penalty details and coefficients are given in Appendix~\ref{app:reward-design-details}.

\paragraph{Policy Optimization.}
For each task $q$ with candidate set $\mathcal{M}(q)$, we sample a group of $G$ memory contexts $\{S_g\}_{g=1}^{G}$ from the current policy $\pi_{\theta_{\mathrm{old}}}$ and obtain their ExPO rewards $\{r_g^{\mathrm{ExPO}}\}_{g=1}^{G}$ using Eq.~\ref{eq:expo-reward}.

Following DAPO~\citep{yu2025dapo}, we discard groups whose task-outcome rewards (Eq.~\ref{eq:task-reward}) are all identical and continue sampling until the batch is filled. 
Filtering on task outcomes rather than on the full reward ensures that every retained group differs in task success, so that the selection-size and format penalties alone cannot make a group informative.

For each retained group, the advantage of a context is its group-centered ExPO reward:
\begin{equation}
  A_g
  =
  r_g^{\mathrm{ExPO}}
  - \frac{1}{G}\sum_{h=1}^{G} r_h^{\mathrm{ExPO}} .
\label{eq:adv}
\end{equation}
Following \citet{liu2025understanding}, we do not normalize advantages by the group standard deviation, which would amplify reward noise in groups with nearly identical outcomes.

ExPO optimizes the clipped surrogate objective of GRPO:
\begin{equation}
\begin{aligned}
  \ell_{g,t}(\theta)
  = \min\Bigl\{&
    \rho_{g,t}(\theta) A_g,\\
    &\operatorname{clip}\bigl(
      \rho_{g,t}(\theta), 1-\epsilon, 1+\epsilon
    \bigr) A_g
  \Bigr\},
\end{aligned}
\end{equation}
where $\rho_{g,t}(\theta)$ is the token-level likelihood ratio between $\pi_\theta$ and $\pi_{\theta_{\mathrm{old}}}$, conditioned on $q$, $\mathcal{M}(q)$, and the preceding output tokens, and $\epsilon$ is the clipping threshold.

The ExPO training loss aggregates the surrogate at the token level over all sequences in a batch:
\begin{equation}
  \mathcal{L}_{\mathrm{ExPO}}(\theta)
  =
  -\,\frac{1}{\sum_{g} T_g}
  \sum_{g}\sum_{t=1}^{T_g}
  \ell_{g,t}(\theta),
\end{equation}
where $T_g$ is the number of generated tokens in the composer output encoding $S_g$, and the sums run over all sequences of all groups in the batch.
ExPO thus inherits the optimization machinery of GRPO, while its advantages are derived from the memory-assembly reward in Eq.~\ref{eq:expo-reward}. ExPO uses no KL penalty to a reference policy, and only \method{} is updated; the executor, retriever, and memory bank remain fixed throughout training. Hyperparameters and other details are provided in Appendix~\ref{app:training-details}.

\begin{table*}[t]
\centering
\small
\renewcommand{\arraystretch}{1.08}
\setlength{\tabcolsep}{7pt}
\begin{tabular}{l cc cc cc cc}
\toprule
\multirow{2}{*}{\textbf{Method}}
& \multicolumn{2}{c}{\textbf{Retail}}
& \multicolumn{2}{c}{\textbf{Airline}}
& \multicolumn{2}{c}{\textbf{Telecom}}
& \multicolumn{2}{c}{\textbf{Avg.}} \\
\cmidrule(lr){2-3}
\cmidrule(lr){4-5}
\cmidrule(lr){6-7}
\cmidrule(lr){8-9}
& pass$^1$ & pass$^4$
& pass$^1$ & pass$^4$
& pass$^1$ & pass$^4$
& pass$^1$ & pass$^4$ \\
\midrule
\multicolumn{9}{l}{\qwenicon \ \textbf{Qwen3-4B-Instruct}} \\
\midrule
No Memory
& 45.6 & 30.0
& 32.5 & 10.0
& 22.5 & 2.5
& 33.5 & 14.2 \\

\midrule
Similarity-based
& 53.7 & 30.0
& \underline{50.0} & 25.0
& 29.4 & 5.0
& 44.4 & 20.0 \\

Complexity-based
& 51.9 & \underline{37.5}
& 40.0 & \underline{30.0}
& \underline{37.5} & \underline{15.0}
& 43.1 & \underline{27.5} \\

\midrule
Qwen3-Reranker-8B
& 54.4 & 35.0
& 46.2 & \underline{30.0}
& 31.9 & 7.5
& 44.2 & 24.2 \\

MemRL
& \underline{56.9} & 22.5
& 41.2 & 10.0
& 35.6 & \textbf{20.0}
& 44.6 & 17.5 \\

Memento
& 53.1 & \underline{37.5}
& 46.2 & \textbf{35.0}
& 35.6 & 7.5
& \underline{45.0} & 26.7 \\

Vanilla Selector
& 50.0 & 22.5
& 43.8 & \underline{30.0}
& 34.4 & 10.0
& 42.7 & 20.8 \\

\rowcolor{blue!12}
\textbf{\method{}}
& \textbf{60.0} & \textbf{42.5}
& \textbf{52.5} & \textbf{35.0}
& \textbf{40.0} & \textbf{20.0}
& \textbf{50.8} & \textbf{32.5} \\

\midrule
\multicolumn{9}{l}{\mistralicon \ \textbf{Ministral-3-8B-Instruct}} \\
\midrule
No Memory
& 58.1 & 32.5
& 17.5 & 10.0
& 51.9 & 25.0
& 42.5 & 22.5 \\

\midrule
Similarity-based
& 68.1 & 40.0
& \underline{38.8} & \textbf{20.0}
& 63.7 & 45.0
& 56.9 & 35.0 \\

Complexity-based
& 65.6 & \underline{45.0}
& 36.2 & \textbf{20.0}
& 73.8 & 40.0
& 58.5 & 35.0 \\

\midrule
Qwen3-Reranker-8B
& 62.5 & 35.0
& 28.7 & 5.0
& 66.2 & 37.5
& 52.5 & 25.8 \\

MemRL
& \underline{70.6} & \underline{45.0}
& 30.0 & \underline{15.0}
& \textbf{76.2} & \textbf{55.0}
& \underline{58.9} & \underline{38.3} \\

Memento
& 66.9 & \textbf{47.5}
& 36.2 & 10.0
& 70.0 & 45.0
& 57.7 & 34.2 \\

Vanilla Selector
& 64.4 & 30.0
& 37.5 & 10.0
& 66.9 & 45.0
& 56.3 & 28.3 \\

\rowcolor{blue!12}
\textbf{\method{}}
& \textbf{75.0} & \textbf{47.5}
& \textbf{40.0} & \textbf{20.0}
& \underline{75.6} & \underline{52.5}
& \textbf{63.5} & \textbf{40.0} \\

\bottomrule
\end{tabular}
\caption{\textbf{Results on $\tau^2$-bench with two frozen executor backbones.} pass$^k$ is the probability that a task is solved in all $k$ independent trials~\citep{yao2024tau}: pass$^1$ is the average success rate and pass$^4$ reflects consistency.
Heuristic baselines use the $k \in \{1, 3, 5, 7\}$ with the highest average pass$^1$ on the test set. 
Avg.\ is the macro average over domains. Bold and underline mark the best and second-best results within each backbone; ties share the marking.}
\label{tab:main}
\end{table*}

\section{Experiments}

\subsection{Datasets and Evaluation Metrics}
We evaluate on $\tau^2$-bench~\citep{barres2025tau2}, a benchmark of multi-turn customer-service tasks in which an agent must follow a domain policy, call tools, and interact with a simulated user. We use the \textsc{retail}, \textsc{airline}, and \textsc{telecom} domains, and the user is simulated by GLM-4.7~\citep{glm45} for all methods. 

Following prior work~\citep{yao2024tau}, we run each test task for four trials and report pass$^1$, the average success rate, as the primary metric, and pass$^4$, the fraction of tasks solved in all four trials, to assess reliability. Averages are taken over the three domains with equal weight. Further details are provided in Appendix~\ref{app:eval}.

\subsection{Baselines}
For each agent backbone, all methods use the same frozen executor, user simulator, and memory bank constructed from that agent's rollouts, and select from the same top-10 candidates. We compare against the following baselines.
\paragraph{No memory.} The executor solves each task without retrieved memories.
\paragraph{Heuristic selection.} Similarity-based selection provides the $k$ candidates whose first user messages are most similar to the current one under Qwen3-Embedding-8B~\citep{zhang2025qwen3embedding}, ordered from most to least similar. 
Complexity-based selection, inspired by complexity-based prompting~\citep{fu2023complexity}, provides the $k$ candidates with the longest trajectories, ordered from longest to shortest. 
For both, we report the best $k \in \{1, 3, 5, 7\}$ on the test set. 
Results for all $k$ are given in Appendix~\ref{app:full-results}.

\paragraph{Model-based selection.} The vanilla selector applies our selection prompt to the agent backbone without training. The pre-trained reranker reranks the candidates with Qwen3-Reranker-8B and provides the top five.

\paragraph{RL-based memory methods.} We compare \method{} with two RL-based methods, MemRL~\citep{zhang2026memrl} and Memento~\citep{zhou2025memento}, which learn the utility of memories from task outcomes. 
Implementation details are given in Appendix~\ref{app:baselines}.

\subsection{Main Results}
Table~\ref{tab:main} shows that \method{} achieves the highest average pass$^1$ and pass$^4$ with both executor backbones, and is the best or tied-best method in 10 of the 12 domain-level comparisons. 
On Qwen3-4B-Instruct, it reaches 50.8 average pass$^1$ and 32.5 pass$^4$, outperforming the strongest baseline on each metric by 5.9 and 5.0 points, respectively. 
On Ministral-3-8B-Instruct, it achieves 4.6 points above the strongest baseline with a pass$^1$ of 63.5, and slightly above the best baseline with a pass$^4$ of 40.0. Notably, no baseline is consistently competitive: Memento is the strongest baseline with Qwen3-4B-Instruct but trails with Ministral-3-8B-Instruct, while MemRL is the strongest with Ministral-3-8B-Instruct but reaches only 17.5 average pass$^4$ with Qwen3-4B-Instruct, close to the 14.2 of the no-memory agent. 


Pooled over both executors, the gap to every baseline is significant at the 95\% level on pass$^1$, and on pass$^4$ for all baselines except Complexity and Memento (paired bootstrap over test tasks; Appendix~\ref{app:ci}). 
Beyond $\tau^2$-bench, \method{} also achieves the highest overall success rate on ALFWorld~\citep{alfworld} with Ministral-3-8B-Instruct, at 82.0\% compared with 79.7\% for the strongest baseline (Appendix~\ref{app:alfworld}).
Finally, \method{} reaches this performance with fewer memory tokens than the strongest baseline, 16.9\% fewer with Qwen3-4B-Instruct and 4.9\% fewer with Ministral-3-8B-Instruct (Appendix~\ref{app:efficiency}).



\section{Findings}
\label{sec:analysis}

\subsection{How Does Training Change Experience Selection?}
We find that training changes selection in three ways: \method{} chooses more experiences from deeper in the ranking, shows a weaker preference for historically successful experiences, and learns to order the selected experiences more effectively than fixed rules.

The first two analyses compare the vanilla selector with \method{} on the same 40 retail tasks and top-10 candidate pools with Qwen3-4B-Instruct as the executor; the ordering analysis uses Ministral-3-8B-Instruct on all test tasks.

\paragraph{Training makes \method{} choose more, and deeper.}
\begin{figure}[t]
  \centering
  \includegraphics[width=\columnwidth]{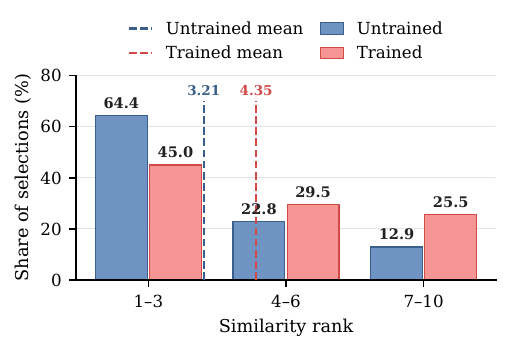}
  \caption{
\textbf{Lower-ranked candidates account for a larger share of selections after training}. Dashed lines indicate the mean rank of selected memories.}
  \label{fig:selection-rank}
\end{figure}

Training increases the average selection size from 2.52 to 3.73 experiences compared to the vanilla selector. After training, \method{} also chooses more experiences from farther down the retrieval ranking (Figure~\ref{fig:selection-rank}), and the mean similarity rank of all selected experiences rises from 3.21 to 4.35. The learned policy draws more broadly from the candidate pool, consistent with our intuition that useful memory contexts may include guidance beyond the closest matches.

\paragraph{Training weakens the preference for successful experiences.}
\begin{figure}[t]
  \centering
  \includegraphics[width=\linewidth]{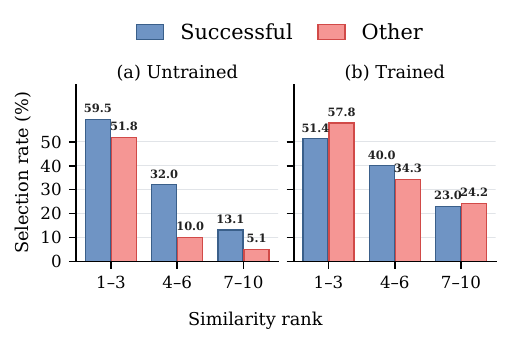}
  \caption{
    \textbf{Training weakens the preference for successful experiences.}
    Fraction of candidates selected within each rank band, for the (a) untrained and (b) trained selectors on retail.
    }
  \label{fig:selection-success}
\end{figure}
Before training, successful experiences are selected more often than failed or fake-pass in every rank band, most clearly at ranks 4--6 (32\% vs.\ 10\%; Figure~\ref{fig:selection-success}). After training, this gap narrows to 6 points at ranks 4--6, reverses at ranks 1--3, and vanishes at ranks 7--10; overall, the share of successful experiences among selections falls from 45.5\% to 35.6\%.

This shift follows from the training objective: \method{} is rewarded for the assembled context's effect on the current task, not for the memories' past outcomes, and a failed memory can still be instructive. For example, a memory that miscalculated a price difference nonetheless demonstrates the exchange operation the new task requires, raising success from 0/4 to 4/4 trials compared with no memory (see Table~\ref{tab:case-failed-retail} for details).

\paragraph{\method{} orders selected memories better than fixed rules.}
\begin{table}[t]
\centering
\small
\setlength{\tabcolsep}{6pt}
\begin{tabular}{lcc}
\toprule
\textbf{Alternative order}
& \textbf{Mean $\Delta$}
& \textbf{$\Delta$ pass$^1$ (points)} \\
\midrule
Similarity         & +0.220 & +5.5 \\
Similarity reversed & +0.340 & +8.5 \\
Length $\uparrow$      & +0.270 & +6.8 \\
Length $\downarrow$    & +0.150 & +3.8 \\
\midrule
\textbf{Average} & \textbf{+0.245} & \textbf{+6.1} \\
\bottomrule
\end{tabular}
\caption{
    \textbf{Advantage of the composer's order} over alternative orderings of the same memories. $\Delta$: mean difference in successful trials over 100 tasks; $\Delta$ pass$^1 = 25\Delta$ points. Positive values favor the composer.
}
\label{tab:ordering-gain}
\end{table}
To isolate the effect of ordering, we keep the memories selected by \method{} for each task fixed and reorder them by one of four rules: similarity, reverse similarity, length descending, or length ascending. 
As Table~\ref{tab:ordering-gain} shows, the composer's order outperforms every rule by 3.8 to 8.5 points and 6.1 on average in pass$^1$; since the memory set is identical, the gain comes solely from order.

Case 3 in Appendix~\ref{app:case} gives an example: the same five memories yield 4/4 successes under the composer's order and 0/4 under similarity order. Assembling a useful memory context therefore requires deciding not only which memories to include but how to present them.

\begin{figure*}[t]
  \centering
  \includegraphics[width=\textwidth]{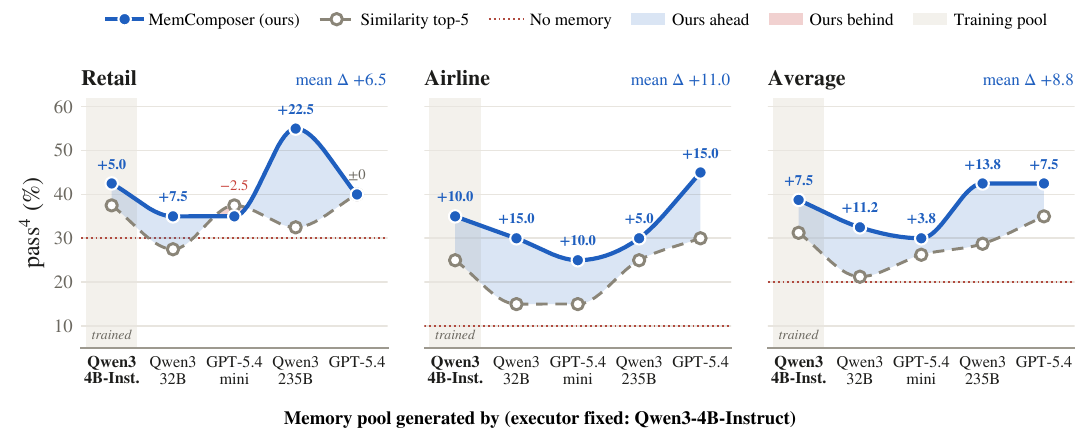}
  \caption{\textbf{Effect of the memory source.}
  Only the agent that generated the memory pool varies along the x-axis. Qwen3-4B-Instruct is the executor's own memory, on which \method{} was trained. Both methods draw from the same top-10 candidates per task:
  similarity top-5 keeps the five most similar, while \method{} selects its own subset.
  We report pass$^4$ on retail (40 tasks), airline (20 tasks), and their unweighted average.
  Shading marks where \method{} is ahead (blue) or behind (red).
  Markers are measured values; curves are a visual guide only.}
\label{fig:memory}
  \label{fig:memory}
\end{figure*}

\subsection{Generalization}
We apply the trained composer without further training to unseen domains, to a different agent backbone, and to memory banks of different quality to test its transferability.

\paragraph{\method{} transfers effectively across domains.}
\begin{table}[t]
    \centering
    \small
    \setlength{\tabcolsep}{4pt}
    \begin{tabular}{l cc cc}
    \toprule
    \multirow{2}{*}{\textbf{Method}} & \multicolumn{2}{c}{\textbf{Airline}} & \multicolumn{2}{c}{\textbf{Telecom}} \\
    \cmidrule(lr){2-3}\cmidrule(lr){4-5}
    & pass$^1$ & pass$^4$ & pass$^1$ & pass$^4$ \\
    \midrule
    No Memory             & 32.5 & 10.0 & 22.5 & 2.5 \\
    Similarity$^\dagger$  & 50.0 & 25.0 & 29.4 & 5.0 \\
    Complexity$^\dagger$  & 40.0 & \underline{30.0} & 37.5 & 15.0 \\
    Vanilla selector      & 43.8 & \underline{30.0} & 34.4 & 10.0 \\
    \midrule
    \multicolumn{5}{l}{\textit{\method{}}} \\
    \rowcolor{blue!12}
    \quad In-domain       & \underline{52.5} & \textbf{35.0} & \textbf{40.0} & \underline{20.0} \\
    \rowcolor{red!9}
    \quad Retail$\rightarrow$ & \textbf{53.8} & \textbf{35.0} & \underline{38.8} & \textbf{22.5} \\
    \bottomrule
    \end{tabular}
    \caption{\textbf{The retail-trained composer transfers to unseen domains.}
    Qwen3-4B-Instruct executor. \emph{In-domain}: trained on the target
    domain; \emph{Retail}$\rightarrow$: trained on retail only.
    $^\dagger$As in Table~\ref{tab:main}.}
    \label{tab:cross-domain}
\end{table}
We apply the composer trained on \textsc{retail} to \textsc{airline} and \textsc{telecom} with the same Qwen3-4B-Instruct executor and each target domain's memory bank, without further training.

As shown in Table~\ref{tab:cross-domain}, the transferred \method{} outperforms the vanilla selector and both heuristics in both domains, and performs comparably to in-domain training: on \textsc{airline} it is on par in pass$^1$ (53.8 vs.\ 52.5) and equal in pass$^4$, and on \textsc{telecom} it trades 1.2 points of pass$^1$ for 2.5 points of pass$^4$. The composer thus captures selection principles that hold across domains with different tools and policies, rather than retail-specific shortcuts. We next ask whether these principles also hold for a different executor.

\paragraph{The learned policy transfers to a different executor.}
\begin{table}[t]
    \centering
    \small
    \setlength{\tabcolsep}{6pt}
    \begin{tabular}{l cc}
    \toprule
    \textbf{Method} & pass$^1$ & pass$^4$ \\
    \midrule
    No Memory                  & 58.1 & 32.5 \\
    Similarity $^\dagger$         & 68.1 & 40.0 \\
    Complexity $^\dagger$         & 65.6 & 45.0 \\
    \midrule
    Vanilla selector & 64.4 & 30.0 \\
    \rowcolor{blue!12}
    \method{}
                               & \textbf{75.0} & \underline{47.5} \\
    \midrule
    \rowcolor{red!9}
    \textbf{\method{}} (4B)
                               & \underline{70.6} & \textbf{50.0} \\
    \bottomrule
    \end{tabular}
    \caption{
    \textbf{The learned policy transfers to a different executor.} Retail, Ministral-3-8B-Instruct executor with its own memory bank.
    \method{}: trained with Ministral-3-8B-Instruct; \method{} (4B): trained with Qwen3-4B-Instruct. $^\dagger$As in Table~\ref{tab:main}.}
    \label{tab:cross-executor}
\end{table}
On retail, we pair the Qwen3-4B-Instruct composer with a frozen Ministral-3-8B-Instruct executor and its self-generated memory, without further composer training. All compared methods use the same target executor and memory.
As shown in Table~\ref{tab:cross-executor}, the transferred composer achieves 70.6 pass$^1$ and 50.0 pass$^4$, outperforming both heuristic baselines and the untrained composer. Compared with the composer trained for Ministral, transfer yields lower pass$^1$ (70.6 versus 75.0) but higher pass$^4$ (50.0 versus 47.5). These results suggest that the learned assembly policy remains useful with a different executor and experience pool.

\paragraph{Learned assembly remains beneficial as memory improves.}
We test whether the advantage of \method{} persists when the memory is built by stronger agents.
We replace the memory bank with banks built by four other agents, from Qwen3-32B to GPT-5.4, using the same construction pipeline (Appendix~\ref{app:memory}) and setting; since stronger agents tend to produce more reliable trajectories and reviews~\citep{zheng2023judging}, these banks span a range of memory quality.
The executor (Qwen3-4B-Instruct), test tasks, user simulator, and retrieval procedure are held fixed; for each bank, all methods share the same top-10 candidate pool, and the composer, trained only on the executor's own bank, is applied without retraining.

Figure~\ref{fig:memory} shows two patterns.
First, better memory alone yields limited gains under similarity retrieval: on \textsc{retail}, similarity top-5 with GPT-5.4's bank performs close to its level with the executor's own bank.
Second, \method{} matches or outperforms similarity top-5 in 9 of the 10 domain--bank pairs and improves average $\text{pass}^4$ on every unseen bank, by 9.1 points on average and by 7.5 points even with GPT-5.4's bank.
Improving what is stored therefore does not remove the need to decide what to recall.

\section{Conclusion}

We introduced \method{}, which jointly selects and orders a variable-size subset of retrieved experiences for a frozen LLM agent. Through Experience Policy Optimization (ExPO), it learns from downstream task outcomes while keeping the executor, retriever, and memory bank fixed. Across three $\tau^2$-bench domains and two executors, \method{} achieves the highest average $\mathrm{pass}^{1}$ and $\mathrm{pass}^{4}$ among evaluated methods with fewer memory tokens than the strongest baseline for each executor. Further analyses show improved selection and ordering, alongside transfer across domains, executors, and memory sources without retraining. These results highlight the value of learning memory assembly from downstream feedback.

\section*{Limitations}

\paragraph{Executor preferences.}
\method{} learns an assembly policy for one frozen executor. We show that this policy transfers to another. Still, we do not characterize \emph{what} different executors prefer in a memory context: whether a stronger model benefits from fewer, longer, or differently ordered experiences, and whether the preference for failed experiences in \S\ref{sec:analysis} is specific to small executors. Both executors in this work have 4B and 8B parameters; it remains open whether a frontier-scale executor that can judge candidate memories on its own would benefit from a separate composer to the same extent.

\paragraph{Ordering.} \textsc{MemComposer} learns presentation order only implicitly from the sequence-level reward. Within a group, sampled sequences usually differ in both content and order, so credit for ordering is entangled with credit for selection, and with $G=8$ samples per prompt drawn from a space of ordered subsets that grows factorially with $k$, the learned order is not guaranteed to be optimal. Table~\ref{tab:ordering-gain} shows that it outperforms four heuristic orderings of the same memories, but we do not compare it against an oracle that searches over all permutations.

\paragraph{Memory type.}
We study procedural memory: complete past trajectories together with their reviews, from which the executor can reuse actions and avoid earlier mistakes. We do not evaluate whether learned assembly helps with other memory types, such as semantic memory (facts and rules) or episodic memory of a particular user's history, where the interaction between selected items and their order may differ.

\bibliography{custom}

\appendix

\section{Extended Related Work}
\label{app:related-works}

\subsection{Memory for LLM Agent}
Agent memory lets LLM agents reuse past experience without updating
parameters~\citep{jiang2025adaptation}, in the form of trajectories and insights
\citep{zhao2024expel, ouyang2026reasoningbank}, workflows and skills
\citep{wang2025awm, wang2023voyager, ouyang2026skillos,
zhang2026usefulmemory}, or general-purpose memory stores
\citep{packer2023memgpt, chhikara2025mem0, xu2025amem}. Most of this
work improves what is stored and how it is organized, while memories
are still recalled by embedding similarity. Recent methods learn which
memories to recall from task outcomes: MemRL \citep{zhang2026memrl}
maintains a utility value for each memory, and Memento
\citep{zhou2025memento} trains a scorer that predicts whether a single
case will help; Memory-R1 \citep{yan2026memoryr1} uses RL to manage
memory operations. These methods score or operate on memories
individually and return a fixed number of them. \method{} instead
treats the assembled context as the unit of decision, choosing which
memories to include, how many, and in what order, and is rewarded for
the outcome of that context as a whole.

\subsection{Learning Context Selection from Downstream Feedback}
Learning which context to provide from downstream feedback has been
studied in both in-context learning and retrieval-augmented generation.
For in-context learning, EPR \citep{rubin2022epr}, UDR \citep{li2023udr},
and LLM-R \citep{wang2024llmr} train demonstration retrievers using the
model's likelihood of the gold output, while later work selects the
demonstration set jointly: CEIL \citep{ye2023ceil} with a determinantal
point process, Se$^2$ \citep{liu2024se2} with sequential beam search,
and \citet{zhang2022active} and RetICL \citep{scarlatos2023reticl} with
reinforcement learning, motivated in part by the sensitivity of
performance to demonstration order \citep{lu2022fantastically}. For
retrieval-augmented generation, beyond LLM-based
reranking~\citep{tian2026corank}, Rank-R1 \citep{zhuang2025rankr1} and
REARANK \citep{zhang2025rearank} train rerankers with reinforcement
learning against relevance labels, search agents and rerankers have been
trained from the generator's output \citep{jiang2025s3, sun2026barrag},
and DynamicRAG \citep{sun2025dynamicrag} learns both the number and the
order of passages from the quality of the generated answer. Beyond
selection, other work structures how retrieved evidence is used, guiding
multi-hop reasoning with a taxonomy~\citep{sun2026tasrrag} or expressing
it as an executable program~\citep{sun2026pyrag}. 
Our setting differs in what is selected and how it is supervised. The candidates
are full agent trajectories with hindsight reviews, whose value lies in
how they steer an executor over many steps rather than in the facts or
answer formats they supply. Prior methods rely on signals that are
cheap and available per query, such as the likelihood of a gold output,
relevance labels, or the quality of one generated answer. We have
neither a gold output nor a gold memory sequence, and an assembled
context can only be scored after a frozen executor completes a
multi-turn interaction with tools and a simulated user, so the reward
is sparse, stochastic, and expensive. Because a single executor run
confounds context quality with the executor's own randomness, ExPO
estimates each context's utility from repeated executor trials within
a GRPO-style group comparison.

\section{Statistical Significance}
\label{app:ci}

Each $\tau^2$-bench domain has only 20--40 test tasks (Table~\ref{tab:eval-stats}), so we quantify the uncertainty of the gaps reported in Table~\ref{tab:main}.
For every baseline we compute the gap $\Delta = \text{\method{}} - \text{baseline}$ in pass$^1$ and pass$^4$ and estimate a 95\% confidence interval by paired bootstrap over test tasks: in each of 10,000 resamples, we draw tasks with replacement within each domain, recompute the gap from each task's outcomes under both methods, take the equal-weight macro average over domains as in Table~\ref{tab:main}, and report the 2.5th and 97.5th percentiles.
Table~\ref{tab:ci-exec} reports this interval separately for each executor.
Table~\ref{tab:ci-pooled} pools both executors; a resampled task then contributes its outcomes under both executors, so the two are never treated as independent samples.
No correction for multiple comparisons is applied, so intervals that barely exclude zero should be read with that in mind.

\begin{table}[t]
\centering\footnotesize
\setlength{\tabcolsep}{2pt}
\begin{tabular}{lcc}
\toprule
Baseline & $\Delta$ pass$^1$ & $\Delta$ pass$^4$ \\
\midrule
\multicolumn{3}{l}{\textit{Qwen3-4B-Instruct}} \\
No memory            & \textbf{+17.3} \ci{9.2}{25.6}  & \textbf{+18.3} \ci{9.2}{28.3} \\
Similarity ($k{=}7$) & \textbf{+6.5} \ci{0.2}{13.1}   & \textbf{+12.5} \ci{3.3}{21.7} \\
Complexity ($k{=}3$) & +7.7 \ci{0.0}{15.4}            & +5.0 \ci{$-$3.3}{13.3} \\
Qwen3-Reranker       & \textbf{+6.6} \ci{0.2}{13.1}   & +8.3 \ci{$-$0.8}{17.5} \\
MemRL                & +6.3 \ci{$-$0.4}{12.7}         & \textbf{+15.0} \ci{5.0}{25.0} \\
Memento              & +5.9 \ci{$-$0.8}{12.7}         & +5.8 \ci{$-$3.3}{15.0} \\
Vanilla selector     & \textbf{+8.1} \ci{1.6}{14.6}   & \textbf{+11.7} \ci{4.2}{19.2} \\
\midrule
\multicolumn{3}{l}{\textit{Ministral-3-8B-Instruct}} \\
No memory            & \textbf{+21.0} \ci{12.3}{30.0} & \textbf{+17.5} \ci{7.5}{28.3} \\
Similarity ($k{=}3$) & \textbf{+6.7} \ci{0.2}{13.3}   & +5.0 \ci{$-$3.3}{13.3} \\
Complexity ($k{=}7$) & +5.0 \ci{$-$2.1}{12.3}         & +5.0 \ci{$-$5.8}{15.8} \\
Qwen3-Reranker       & \textbf{+11.1} \ci{5.2}{17.3}  & \textbf{+14.2} \ci{5.0}{24.2} \\
MemRL                & +4.6 \ci{$-$1.0}{10.4}         & +1.7 \ci{$-$7.5}{10.8} \\
Memento              & +5.8 \ci{$-$1.3}{12.9}         & +5.8 \ci{$-$5.0}{16.7} \\
Vanilla selector     & \textbf{+7.3} \ci{0.2}{14.1}   & \textbf{+11.7} \ci{0.8}{22.5} \\
\bottomrule
\end{tabular}
\caption{Gap between \method{} and each baseline per executor, averaged over the three domains, with 95\% paired-bootstrap CIs over test tasks in brackets. Bold marks intervals that exclude zero. Heuristic baselines use the budget $k$ reported in Table~\ref{tab:main}.}
\label{tab:ci-exec}
\end{table}

\begin{table}[t]
\centering\footnotesize
\setlength{\tabcolsep}{2pt}
\begin{tabular}{lcc}
\toprule
Baseline & $\Delta$ pass$^1$ & $\Delta$ pass$^4$ \\
\midrule
No memory            & \textbf{+19.2} \ci{13.2}{25.2} & \textbf{+17.9} \ci{10.8}{25.0} \\
Similarity$^\dagger$ & \textbf{+6.6} \ci{2.0}{11.3}   & \textbf{+8.8} \ci{2.5}{15.0} \\
Complexity$^\dagger$ & \textbf{+6.4} \ci{1.1}{11.6}   & +5.0 \ci{$-$2.1}{12.1} \\
Qwen3-Reranker       & \textbf{+8.8} \ci{4.5}{13.2}   & \textbf{+11.2} \ci{5.0}{17.5} \\
MemRL                & \textbf{+5.4} \ci{1.0}{9.9}    & \textbf{+8.3} \ci{1.7}{15.4} \\
Memento              & \textbf{+5.8} \ci{1.0}{10.8}   & +5.8 \ci{$-$1.2}{12.9} \\
Vanilla selector     & \textbf{+7.7} \ci{2.9}{12.3}   & \textbf{+11.7} \ci{5.4}{18.3} \\
\bottomrule
\end{tabular}
\caption{Gap between \method{} and each baseline pooled over both executors (six domain--executor cells, equal weight), with 95\% paired-bootstrap CIs in brackets. Bold marks intervals that exclude zero. $^\dagger$Each executor uses its budget $k$ from Table~\ref{tab:main}.}
\label{tab:ci-pooled}
\end{table}

\paragraph{Discussion.}
Pooled over both executors (Table~\ref{tab:ci-pooled}), the gap to every baseline excludes zero on pass$^1$, and on pass$^4$ for all baselines except Complexity and Memento.
Within a single executor (Table~\ref{tab:ci-exec}), all 28 gaps are positive, but the intervals are roughly 1.5 times wider: on pass$^1$, each executor's gap excludes zero for four of the seven baselines (no memory, similarity, the reranker, and the vanilla selector) and includes zero for Complexity, MemRL, and Memento, including the strongest baseline for each executor (Memento, +5.9, for Qwen3-4B-Instruct; MemRL, +4.6, for Ministral-3-8B-Instruct).
We therefore read Table~\ref{tab:main} as showing a consistent advantage whose magnitude against the strongest individual baseline is not resolved by the official test splits, rather than a decisive per-executor win.

Two aspects of the comparison do not depend on this uncertainty.
First, \method{} is the best or tied-best method in 10 of the 12 domain-level comparisons (Table~\ref{tab:main}).
Second, the memory-token cost of each method is determined by its selections and is not subject to trial randomness: \method{} reaches its performance with 16.9\% fewer memory tokens than the strongest baseline for Qwen3-4B-Instruct and 4.9\% fewer for Ministral-3-8B-Instruct (Appendix~\ref{app:efficiency}).
Even where the performance gap is within its interval, \method{} thus matches the strongest baseline at lower cost.

\section{Additional Result on ALFWorld Benchmark}
\label{app:alfworld}
\begin{table}[t]
\centering
\small
\setlength{\tabcolsep}{3pt}
\resizebox{\columnwidth}{!}{%
\begin{tabular}{l cccccc c}
\toprule
\textbf{Method} & Pick & Look & Clean & Heat & Cool & Pick2 & \textbf{All} \\
\midrule
No Memory             & 56.5 & 44.4 & 19.4 & 8.7  & 9.5  & 17.6 & 25.6 \\
\midrule
Similarity ($k{=}5$)  & \textbf{91.3} & \underline{88.9} & \underline{87.1} & 65.2 & 90.5 & \underline{23.5} & 76.7 \\
Complexity ($k{=}5$)  & 69.6 & 72.2 & 71.0 & \textbf{95.7} & 85.7 & 5.9 & 69.2 \\
\midrule
MemRL ($k{=}7$)       & 82.6 & 77.8 & \textbf{93.5} & 82.6 & \underline{95.2} & 11.8 & 77.4 \\
Memento ($k{=}5$)     & \underline{87.0} & \underline{88.9} & \underline{87.1} & 82.6 & 90.5 & \textbf{29.4} & \underline{79.7} \\
Vanilla Selector      & 82.6 & \textbf{94.4} & 74.2 & 78.3 & \underline{95.2} & 17.6 & 75.2 \\
\rowcolor{blue!12}
\textbf{\method{}}    & \textbf{91.3} & \underline{88.9} & \underline{87.1} & \underline{87.0} & \textbf{100.0} & \underline{23.5} & \textbf{82.0} \\
\bottomrule
\end{tabular}}
\caption{\textbf{Results on ALFWorld.} Ministral-3-8B-Instruct serves as both the frozen executor and the backbone of the vanilla selector and \method{}. 
We report success rates (\%) by task type. Bold and underline mark the best and second-best results; ties share the marking.}
\label{tab:alfworld}
\end{table}
We further evaluate \method{} on ALFWorld~\citep{alfworld}, whose tasks fall into six types (pick and place, look at an object under a lamp, clean, heat, cool, and pick two objects). Following the evaluation setup of SkillRL~\citep{xia2026skillrl}, we report the success rate on the 133 games of the unseen test split. 
Ministral-3-8B-Instruct serves as both the frozen executor and the backbone of the vanilla selector and \method{}. 
The memory bank is built from the executor's own trajectories on the training games with the same pipeline as for $\tau^2$-bench (Appendix~\ref{app:memory}), and the composer is trained with ExPO using task success as the outcome. 
All baselines draw from the same top-10 candidates, and fixed-budget baselines
use $k=5$, and MemRL uses $k=7$. 
For \method{}, we report the checkpoint after 20 training steps, roughly one epoch over the training data.

As shown in Table~\ref{tab:alfworld}, memory is highly beneficial in this environment: similarity-based selection alone raises the success rate from 25.6\% to 76.7\%. \method{} achieves the highest overall success rate, 82.0\%, compared with 79.7\% for Memento, the strongest baseline, and is best or tied-best on three of the six task types. 
The most difficult type, picking two objects, remains challenging for all methods, with no method exceeding 30\%.

\section{Memory Bank Construction}
\label{app:memory}

\begin{table}[t]
\centering
\small
\setlength{\tabcolsep}{4pt}
\resizebox{\columnwidth}{!}{%
\begin{tabular}{ll rrrr}
\toprule
\textbf{Agent} & \textbf{Domain} & \textbf{\#} & \textbf{Clean} & \textbf{Fake} & \textbf{Fail} \\
\midrule
\multirow{3}{*}{Qwen3-4B}
  & \textsc{retail}  & 143 & 37.8 & 2.8  & 59.4 \\
  & \textsc{airline} & 58  & 19.0 & 19.0 & 62.1 \\
  & \textsc{telecom} & 141 & 12.8 & 8.5  & 78.7 \\
\midrule
\multirow{3}{*}{Ministral-3-8B}
  & \textsc{retail}  & 144 & 27.1 & 11.1 & 61.8 \\
  & \textsc{airline} & 57  & 15.8 & 14.0 & 70.2 \\
  & \textsc{telecom} & 144 & 22.9 & 6.9  & 70.1 \\
\bottomrule
\end{tabular}}
\caption{Memory bank statistics: number of memories (\#) and the share (\%)
of clean passes, fake passes, and failures.}
\label{tab:memory-stats}
\end{table}

This section describes how the memory bank used by \method{} and all baselines is constructed. The same procedure is applied to every agent. For each agent, the memory bank is built only from that agent's own rollouts on the training split, so no test task contributes to memory. Table~\ref{tab:memory-stats} summarizes the resulting memory banks.

\paragraph{Rollout collection.}
For each training task in a domain, the frozen agent performs $N=8$ independent rollouts against the user simulator (GLM-4.7), following the standard $\tau^2$-bench protocol. Each rollout is scored by the environment's outcome reward, which indicates whether the final database state and the agent's communicated information satisfy the task requirements. We observed no simulator or environment failures during collection, so every rollout is passed to the review stage.

\paragraph{Self-review \& Pass-audit.}
Each rollout is then reviewed in hindsight by the same model that produced it. Because reviews are written only for training tasks, the reviewer is given the task outcome, the evaluator's feedback, and the reference solution.
For a \emph{failed} rollout, the reviewer identifies the first agent turn at which the trajectory went wrong and states what the agent should have done instead; when another rollout of the same task succeeded, it is shown as a
contrastive example. For a \emph{successful} rollout, the reviewer checks whether the agent made critical errors along the way, such as giving the user false information or claiming an action it never performed. The outcome reward itself is never changed: a successful rollout in which the reviewer finds a critical error is labeled \emph{fake pass}, while other successful rollouts are labeled \emph{clean pass}, and failed rollouts are labeled \emph{failure}. The label and the review are stored alongside the trajectory. Prompts are provided in Appendix~\ref{app:prompts}.

\paragraph{Consistency Check.}
Each review is validated by deterministic consistency checks before it is stored. A review must cite a turn that exists in the trajectory and is produced by the agent, its claimed outcome must match the environment's reward, and its output must follow the required format. Reviews that fail any check are discarded together with their rollout.

\paragraph{Curation.}
To keep the memory compact and balanced, we retain at most two entries per training task: one clean or fake pass and one failure when both exist, and otherwise two rollouts with the same outcome chosen at random.

\paragraph{Representation.}
A full trajectory with its review averages 2.3K--3.0K tokens, so ten candidates would not fit comfortably in a selector's context. We therefore render each entry in two forms.  The \emph{digest} (Memory~\ref{mem:digest}), shown to \method{} and the other model-based selectors, contains the user's first message, the outcome label, a short summary of the agent's key actions, and the review, and averages 1.0K--1.4K tokens. The \emph{full form} (Memory~\ref{mem:full}), provided to the agent once an entry is selected, contains the complete trajectory followed by the review.

\paragraph{Retrieval.}
\label{app:memory-retreival}
Entries are indexed by the user's first message, the only task information available to the agent at the start of an episode, and embedded with Qwen3-Embedding-8B~\citep{zhang2025qwen3embedding}. For each task, the $n=10$ entries with the highest cosine similarity form the candidate pool shared by all methods. During training, entries originating from the same training task are excluded from its candidates.

\section{\method{} Implementation Details}
\label{app:method}

\subsection{Input and Output Format}
\label{app:method-io}

As illustrated in Example~\ref{ex:composer-input},
\method{} receives task instructions, ten retrieved memory digests, the user's opening request,
and output-format instructions. Candidates are numbered \texttt{[1]}--\texttt{[10]} in retrieval order, and each digest summarizes a past interaction, its outcome, and its hindsight review. The input and reasoning are abbreviated for readability and the full prompt template is provided in Appendix~\ref{app:prompts}.

\begin{examplebox}[label={ex:composer-input}]{\method{} Input (Excerpt)}
You are the memory selector for the retail agent.
...
Return the selected case indices in descending order of usefulness. If no retrieved case provides useful transferable guidance, return an empty list.

Retrieved cases:
[1]: Task: You want to exchange the water bottle and the desk lamp ...
KIND: failure_attribution
FAILED CONDITION: The agent performed an exchange with wrong item IDs ...
SUMMARY: ...
Chat History: ...
ROOT CAUSE: ...
SHOULD HAVE: ...

[2]: ...
...
[10]: ...

Query:
Hi, I'd like to exchange a water bottle and a desk lamp from a recent order.

First, inside <reasoning> </reasoning>, briefly reason about which cases you keep and why. Then output a JSON list of case numbers inside <answer> </answer>.
\end{examplebox}

The composer outputs reasoning followed by an ordered list of distinct candidate indices, dynamically determining both the selection size and presentation order. In Example~\ref{ex:composer-output}, it selects four memories, which the executor receives in the order $1 \rightarrow 6 \rightarrow 5 \rightarrow 8$. An empty list specifies execution without memory.

\begin{examplebox}[label={ex:composer-output}, colbacktitle=teal!30]{\method{} Output (Excerpt)}
<reasoning>
The query involves exchanging a water bottle and a desk lamp ...
</reasoning>
<answer>[1, 6, 5, 8]</answer>
\end{examplebox}

\subsection{Reward Design Details}
\label{app:reward-design-details}
This subsection specifies how the ExPO reward in Eq.~\ref{eq:expo-reward} is computed.

\paragraph{Task-outcome reward.}
Each valid memory sequence is evaluated over $J=8$ trials with the frozen executor. The task, selected memories, and presentation order remain fixed, and trials differ only in their benchmark seeds. We average the binary success labels to obtain $r_g^{\mathrm{task}}$. 
Trials that fail because of infrastructure errors, such as API failures, are resampled, so every sequence is evaluated over $J = 8$ completed trials. When several sequences in a group select the same memories in the same order, their trials are shared.

\paragraph{Format penalty.}
The composer is expected to produce a reasoning section and a list of candidate indices within the answer section (Example~\ref{ex:composer-output}).
An output is invalid if it lacks a closed answer section, contains no list in that section, or includes an index outside $\{1,\ldots,n\}$. 
Invalid outputs are not passed to the executor and receive a final reward of zero.
Repeated indices are de-duplicated, keeping their first occurrence, and an empty list is valid and specifies execution without memory. For valid outputs, a missing reasoning section incurs a penalty of $\lambda_{\mathrm{fmt}}=0.05$, regardless of the task outcome.

\paragraph{Selection-size penalty.}
We apply a penalty only when the selection size $k_g=|S_g|$ exceeds a free allowance $K_{\max}$:
\begin{equation}
  p_{\mathrm{len}}(k_g) = \lambda_{\mathrm{len}} \max(0, k_g - K_{\max}).
\end{equation}
For \textsc{telecom}, we use $\lambda_{\mathrm{len}} = 0.02$ and $K_{\max} = 7$ to discourage the excessive selection sizes observed in preliminary runs. For \textsc{retail} and \textsc{airline}, we set $\lambda_{\mathrm{len}} = 0.01$ and $K_{\max} = 0$.

The penalty is computed for every valid output, independently of its task outcome, and measures memory count rather than token length. With $n=10$, the size penalty is at most $0.1$, below the task-reward increment of $1/8=0.125$. It therefore never reverses the ranking of sequences with different task outcomes and only favors smaller selections among sequences with equal task outcomes and format penalties.

\paragraph{Dynamic sampling.}
Groups are retained only if their task-outcome rewards are not all identical (\S\ref{sec:training}). Differences in the format and size penalties alone therefore never cause a group to be retained. Advantages are then computed from the final rewards as in Eq.~\ref{eq:adv}. Further training settings are given in Appendix~\ref{app:training-details}.

\subsection{Training}
\label{app:training-details}

\paragraph{Training data.}
Training uses the official train split of each domain. Since the agent's first message in $\tau^2$-bench is a fixed greeting, we sample four customer openings per task directly from the user simulator (GLM-4.7, temperature 1.0), which serve as four query variants of the same task. Each opening is embedded with Qwen3-Embedding-8B and used to retrieve memories by cosine similarity, excluding all memories from the query's own task and keeping at most two memories per source task. The top 20 candidates are split into two windows of ten, except for Ministral telecom and airline, which use the top 10 only, and the candidates in each window are renumbered from 1 to 10, so that the selector cannot tell which window it sees and must judge candidates by their content. Each training example pairs one opening with one window. The data carry no labels of which memories are useful: the reward is obtained by running the agent in simulation. Test examples are built with the same pipeline, except that each test task uses a single fixed opening and only the top-10 candidates.

\paragraph{Hyperparameters.}
\begin{table}[t]
\centering
\small
\setlength{\tabcolsep}{4pt}
\begin{tabular}{lc}
\toprule
\multicolumn{2}{c}{\textbf{Qwen3-4B-Instruct}} \\
\midrule
\textbf{Parameter} & \textbf{Value} \\
\midrule
\multicolumn{2}{l}{\textit{LoRA}} \\
\quad Rank / scaling factor       & 64 / 128 \\
\quad Dropout                     & 0 \\
\midrule
\multicolumn{2}{l}{\textit{Rollout sampling}} \\
\quad Candidates per prompt       & 10 \\
\quad Sequences per prompt $G$    & 8 \\
\quad Executor trials per sequence $J$ & 8 \\
\quad Temperature                 & 1.0 \\
\quad Top-$p$ / top-$k$           & 0.9 / 20 \\
\quad Maximum prompt tokens       & 16,384 \\
\quad Maximum output tokens       & 1,024 \\
\midrule
\multicolumn{2}{l}{\textit{Optimization}} \\
\quad Prompts per generation round & 16 \\
\quad Retained groups per update   & 8 \\
\quad Optimizer                    & AdamW \\
\quad Learning rate                & $2\times10^{-5}$ \\
\quad Schedule / warmup            & Constant / none \\
\quad Weight decay                 & 0.01 \\
\quad Gradient clipping            & 1.0 \\
\quad Policy clipping $\epsilon$   & 0.2 \\
\quad Dual-clip coefficient        & 3.0 \\
\quad KL / entropy coefficients    & 0 / 0 \\
\quad Advantage normalization      & Mean-centering only \\
\midrule
\multicolumn{2}{l}{\textit{Implementation}} \\
\quad Training framework           & verl 0.7.1 \\
\quad Precision                    & bfloat16 \\
\quad Data-loader seed             & 42 \\
\quad Policy updates               & 30 \\

\bottomrule
\end{tabular}
\caption{Training hyperparameters for the
Qwen3-4B-Instruct composer.}
\label{tab:qwen-train-hparams}
\end{table}
\begin{table}[t]
\centering
\small
\setlength{\tabcolsep}{4pt}
\begin{tabular}{lc}
\toprule
\multicolumn{2}{c}{\textbf{Ministral-3-8B-Instruct}} \\
\midrule
\textbf{Parameter} & \textbf{Value} \\
\midrule
\multicolumn{2}{l}{\textit{LoRA}} \\
\quad Rank / scaling factor       & 64 / 128 \\
\quad Dropout                     & 0 \\
\midrule
\multicolumn{2}{l}{\textit{Rollout sampling}} \\
\quad Candidates per prompt       & 10 \\
\quad Sequences per prompt $G$    & 8 \\
\quad Executor trials per sequence $J$ & 8 \\
\quad Temperature                 & 1.0 \\
\quad Top-$p$ / top-$k$           & 0.9 / 20 \\
\quad Maximum prompt tokens       & 24,576 \\
\quad Maximum output tokens       & 1,024 \\
\midrule
\multicolumn{2}{l}{\textit{Optimization}} \\
\quad Prompts per generation round & 16 \\
\quad Retained groups per update   & 8 \\
\quad Optimizer                    & AdamW \\
\quad Learning rate                & $1\times10^{-5}$ \\
\quad Schedule / warmup            & Constant / none \\
\quad Weight decay                 & 0.01 \\
\quad Gradient clipping            & 1.0 \\
\quad Policy clipping $\epsilon$   & 0.2 \\
\quad Dual-clip coefficient        & 3.0 \\
\quad KL / entropy coefficients    & 0 / 0 \\
\quad Advantage normalization      & Mean-centering only \\
\midrule
\multicolumn{2}{l}{\textit{Implementation}} \\
\quad Training framework           & verl 0.9.1 \\
\quad Precision                    & bfloat16 \\
\quad Data-loader seed             & 42 \\
\quad Policy updates               & 30 \\
\bottomrule
\end{tabular}
\caption{Training hyperparameters for the
Ministral-3-8B-Instruct composer.}
\label{tab:ministral-train-hparams}
\end{table}

We use verl \citep{verl} for training, and the hyperparameters used for both models are provided in tables~\ref {tab:qwen-train-hparams} and~\ref{tab:ministral-train-hparams}, respectively. We train for 30 policy updates and evaluate the final checkpoint.

\paragraph{Training cost.}
We train each composer for 30 policy updates on NVIDIA H100 80GB GPUs, and report GPU-hours summed over all GPUs used by a run.
For Qwen3-4B-Instruct, one GPU runs composer training and sampling, and five (\textsc{retail}, \textsc{airline}) or seven (\textsc{telecom}) GPUs serve the executor with vLLM, for a total of 24.2, 16.3, and 81.7 GPU-hours on \textsc{retail}, \textsc{airline}, and \textsc{telecom}.
For Ministral-3-8B-Instruct, two GPUs run the composer and the executor is accessed through the AWS Bedrock API, for a total of 7.1, 16.2, and 24.9 GPU-hours on the same domains.
The user simulator is accessed through an API in all runs, so these figures count local compute only.
Most of the cost is environment simulation rather than learning: rollout generation and simulation take 69--91\% of the wall-clock time per training step, while updating a LoRA-adapted 4B or 8B model is inexpensive.
\textsc{telecom} is the most expensive domain in both settings because its conversations are the longest.
The Ministral \textsc{airline} run costs more than \textsc{retail} despite its smaller task pool because identical sequences within a group are rarer, so fewer trials can be shared.

\section{Full Results}
\label{app:full-results}
Table~\ref{tab:full-sweep} reports similarity- and complexity-based selection for $k \in \{1, 2, 3, 5, 7\}$.

For each domain and executor, both methods use the same memory bank and evaluation protocol as the main experiments. We report pass$^1$ and pass$^4$ for each domain, together with their macro-averages across the three domains. Performance varies non-monotonically with $k$, and the best budget depends on both the domain and the executor. 

\begin{table*}[t]
\centering
\small
\setlength{\tabcolsep}{5pt}
\begin{tabular}{l cc cc cc cc}
\toprule
\multirow{2}{*}{\textbf{Method}}
& \multicolumn{2}{c}{\textbf{Retail}}
& \multicolumn{2}{c}{\textbf{Airline}}
& \multicolumn{2}{c}{\textbf{Telecom}}
& \multicolumn{2}{c}{\textbf{Avg.}} \\
\cmidrule(lr){2-3} \cmidrule(lr){4-5} \cmidrule(lr){6-7} \cmidrule(lr){8-9}
& pass$^1$ & pass$^4$ & pass$^1$ & pass$^4$ & pass$^1$ & pass$^4$ & pass$^1$ & pass$^4$ \\
\midrule
\multicolumn{9}{l}{\qwenicon \ \textbf{Qwen3-4B-Instruct}} \\
\midrule
\multicolumn{9}{l}{\textit{Similarity-based}} \\
\quad $k=1$  & 49.4 & 30.0 & 43.8 & 20.0 & 25.6 & 10.0 & 39.6 & 20.0 \\
\quad $k=2$  & 46.3 & 27.5 & 40.0 & 20.0 & 36.2 & 12.5 & 40.8 & 20.0 \\
\quad $k=3$  & 51.2 & 42.5 & 38.8 & 15.0 & 37.5 & 20.0 & 42.5 & 25.8 \\
\quad $k=5$  & 56.2 & 37.5 & 45.0 & 25.0 & 28.1 & 7.5 & 43.1 & 23.3 \\
\quad $k=7$  & 53.7 & 30.0 & 50.0 & 25.0 & 29.4 & 5.0 & 44.4 & 20.0 \\
\midrule
\multicolumn{9}{l}{\textit{Complexity-based}} \\
\quad $k=1$  & 51.2 & 35.0 & 38.8 & 25.0 & 26.3 & 5.0 & 38.8 & 21.7 \\
\quad $k=2$  & 50.6 & 30.0 & 41.2 & 20.0 & 32.5 & 10.0 & 41.5 & 20.0 \\
\quad $k=3$  & 51.9 & 37.5 & 40.0 & 30.0 & 37.5 & 15.0 & 43.1 & 27.5 \\
\quad $k=5$  & 50.0 & 32.5 & 48.7 & 30.0 & 28.1 & 10.0 & 42.3 & 24.2 \\
\quad $k=7$  & 55.0 & 35.0 & 41.2 & 25.0 & 27.5 & 7.5 & 41.2 & 22.5 \\
\midrule
\multicolumn{9}{l}{\mistralicon \ \textbf{Ministral-3-8B-Instruct}} \\
\midrule
\multicolumn{9}{l}{\textit{Similarity-based}} \\
\quad $k=1$  & 58.1 & 45.0 & 28.7 & 5.0 & 59.4 & 32.5 & 48.7 & 27.5 \\
\quad $k=2$  & 63.1 & 35.0 & 37.5 & 15.0 & 57.5 & 37.5 & 52.7 & 29.2 \\
\quad $k=3$  & 68.1 & 40.0 & 38.8 & 20.0 & 63.7 & 45.0 & 56.9 & 35.0 \\
\quad $k=5$  & 65.0 & 45.0 & 30.0 & 5.0 & 70.6 & 45.0 & 55.2 & 31.7 \\
\quad $k=7$  & 65.6 & 45.0 & 31.2 & 5.0 & 69.4 & 40.0 & 55.4 & 30.0 \\
\midrule
\multicolumn{9}{l}{\textit{Complexity-based}} \\
\quad $k=1$  & 59.4 & 30.0 & 28.7 & 5.0 & 60.0 & 35.0 & 49.4 & 23.3 \\
\quad $k=2$  & 61.3 & 37.5 & 28.7 & 10.0 & 61.3 & 27.5 & 50.4 & 25.0 \\
\quad $k=3$  & 65.0 & 37.5 & 31.2 & 5.0 & 70.6 & 47.5 & 55.6 & 30.0 \\
\quad $k=5$  & 61.3 & 40.0 & 30.0 & 5.0 & 70.0 & 45.0 & 53.8 & 30.0 \\
\quad $k=7$  & 65.6 & 45.0 & 36.2 & 20.0 & 73.8 & 40.0 & 58.5 & 35.0 \\
\bottomrule
\end{tabular}
\caption{Full retrieval-budget sweep for similarity-based and complexity-based reranking with two frozen executors. Avg.\ denotes the macro-average across the three domains.}
\label{tab:full-sweep}
\end{table*}

\begin{figure*}[t]
  \centering
  \includegraphics[width=\textwidth]{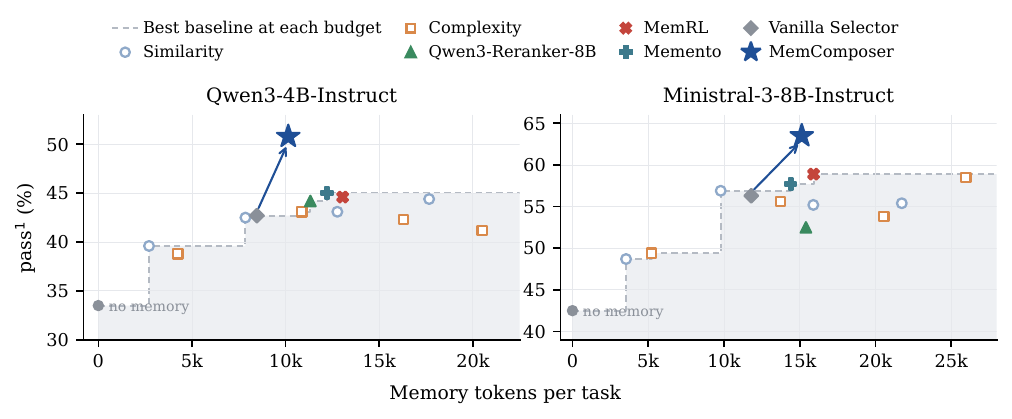}
  \caption{
    Performance--memory-token trade-offs across 2 executor backbones and all domains. Each point represents a method or retrieval-budget configuration. The dashed line shows the baseline Pareto frontier: the highest observed baseline pass$^1$ within each memory-token budget. Memory tokens count the selected context supplied to the executor, averaged per task.
    }
  \label{fig:efficiency-all}
\end{figure*}

\section{Baseline Implementation Details}
\label{app:baselines}
This section describes the implementation and adaptation of all baselines. Every method draws from the same pool of retrieved candidates for each task (Appendix~\ref{app:memory-retreival}), so that differences between methods are attributable to how memories are selected rather than to retrieval.

\subsection{Heuristic Baselines}

\paragraph{Similarity-based selection.}
The $k$ most similar of the top-10 retrieved candidates are provided to the agent, ordered from most to least similar. We sweep $k \in \{1, 2, 3, 5, 7\}$; Table~\ref{tab:main} reports the best $k \in \{1, 3, 5, 7\}$, and results for $k$ are given in Appendix~\ref{app:full-results}.

\paragraph{Complexity-based selection.}
Inspired by complexity-based prompting~\citep{fu2023complexity}, the top-10 candidates are reranked by the length of their rendered full form, measured in tokens, as a cheap proxy for how much procedural detail a trajectory carries, and the $k$ longest are provided to the agent, longest first. 
Because the reranking is applied to the whole pool before truncation, for $k<10$ this baseline selects a different \emph{set} of memories than similarity-based selection, not merely a different order, making it a selection heuristic in its own right. 
At $k=10$, both baselines provide the same memories and differ only in their order. We sweep the same values of $k$ as above.

\subsection{Off-the-Shelf Reranker}

\paragraph{Qwen3-Reranker-8B.}
We rerank the top-10 candidates with Qwen3-Reranker-8B~\citep{zhang2025qwen3embedding}, a cross-encoder that scores the relevance of each candidate to the query.
The query is the user's first message, and each document is the candidate's digest---the same rendering our \method{} reads, so the two methods see identical content and differ only in how they score it. Following the model's recommended usage, the candidate is scored by the probability of ``yes'' at the final position, normalized over the \{``yes'', ``no''\} pair alone rather than the full vocabulary, conditioned on a task instruction, the query and the document. The five highest-scoring candidates are given to the agent in descending order of score. The reranker is used off the shelf, without further training.

\subsection{RL-Based Memory Methods}

\paragraph{MemRL.}
MemRL~\citep{zhang2026memrl} combines a growing memory bank with a value-aware retrieval policy: each memory carries a scalar value $Q$ that estimates how useful it has been, and retrieval proceeds in two phases, a similarity phase that proposes a broad candidate set followed by a value phase that keeps the highest-$Q$ entries. 
Since our memory bank is fixed, we disable memory write-back and reimplement only the value-aware retrieval, the component comparable to our method.

\paragraph{Memento.}
Memento~\citep{zhou2025memento} replaces similarity retrieval with a learned case-selection policy: a scorer trained on the agent's own episode outcomes predicts whether a stored case will help on a new query, and the highest-scoring cases are retrieved. We adapt the parametric variant from the authors' released code. The scorer encodes the query and the case separately with SimCSE-RoBERTa-base, concatenates their \texttt{[CLS]} vectors (case first, as in the reference), and applies a two-layer head ($2 \times 768 \rightarrow 512 \rightarrow 2$, ReLU, dropout $0.2$) that predicts whether the case is helpful. At inference, the five candidates with the highest predicted probability of being helpful are selected from the same top-10 pool that every other method receives and provided in descending order of that probability. Each case is represented by its digest, so the scorer reads the same rendering as our composer.

\textbf{Training data.} The reference implementation accumulates training examples online as the agent runs. Since our memory bank is fixed, we collect the equivalent signal offline from logged episodes: each example is a (query, case, outcome) triple for a case that was in the agent's context during an episode. 
We use only episodes on training tasks and exclude all episodes on test tasks. 
Repeated (query, case) pairs are merged by summing their successes and trials, and each merged pair is weighted by its trial count in the loss, which is equivalent to one hard-labeled example per episode at a fraction of the cost. This yields 797--3,908 unique pairs per agent--domain cell, drawn from 1.2K--233K (episode, case) observations. In some cells, this is more supervision than the original online procedure would accumulate over the same tasks, so the comparison, if anything, favors Memento.

\textbf{Optimization.} We follow the released hyperparameters: AdamW with learning rate $2\times10^{-5}$, weight decay 0.01 (excluding biases and LayerNorm), a linear schedule with 6\% warmup, gradient clipping at 1.0, batch size 32, three epochs, and the reference's default class reweighting $w_{+} = \max(1, n_{-}/n_{+})$, which ranges from 1.00 to 1.80 across cells. We make two departures: we use a maximum sequence length of 512 instead of 256, since 256 truncates 30--60\% of our digests, and we split training and validation data by query rather than by example, so that no query appears in both.

\section{Evaluation Protocol}
\label{app:eval}

\paragraph{Benchmark and splits.}
We evaluate on the retail, airline, and telecom domains of $\tau^2$-bench~\citep{barres2025tau2}, using the official train/test split of each domain. Memory banks are built only from training tasks (Appendix~\ref{app:memory}), and all results are reported on the test tasks: 40 tasks for retail, 20 for airline, and 40 for telecom (Table~\ref{tab:eval-stats}).

\paragraph{Agents and user simulator.}
The agent is either Qwen3-4B-Instruct-2507 (referred to as Qwen3-4B-Instruct in the main text) or Ministral-3-8B-Instruct, used without any parameter updates and decoded with temperature 0. We serve Qwen3-4B-Instruct locally with vLLM and access Ministral-3-8B-Instruct through the AWS Bedrock API. The user is simulated by GLM-4.7 with temperature 0 for all methods, including memory construction, composer training, and evaluation, so that every method interacts with the same user. Each episode ends when the task is completed, the user ends the conversation, or the agent reaches 50 steps. The selected memories are inserted into the agent's system prompt in the order produced by the selection method (Appendix~\ref{app:prompts}).

\paragraph{Selection at test time.}
For every method that selects from the retrieved candidates, the selection is made once per test task and reused across all trials of that task, so that differences between trials reflect the interaction rather than the selected candidates. \method{} and the vanilla selector decode their selections greedily.

\paragraph{Trials and metrics.}
Apart from the choice of user simulator, we follow the official $\tau^2$-bench evaluation protocol: each test task is run for four trials with the official seeds, and performance is measured by pass$^k$, the metric introduced in $\tau$-bench~\citep{yao2024tau} and adopted by $\tau^2$-bench, which is the probability that the agent succeeds in all $k$ of $k$ independent trials of a task. 
Following the official implementation, for a task with $c$ successes out of four trials, pass$^k$ is estimated as $\binom{c}{k} / \binom{4}{k}$ and averaged over tasks; pass$^1$ thus equals the average success rate, and pass$^4$ the fraction of tasks solved in all four trials. When averaging across domains, we take the unweighted mean of the three domains.

\paragraph{Memory tokens.}
For the efficiency analysis, we measure the number of tokens in the memory block inserted into the agent's prompt, counted with the agent's own tokenizer and including the block's fixed header of 172 tokens; a task without selected memories counts as zero. 
Values are averaged over the test tasks of each domain, and over the three domains weighted by their number of test trials.

\begin{table}[t]
\centering
\small
\begin{tabular}{l ccc}
\toprule
 & \textsc{retail} & \textsc{airline} & \textsc{telecom} \\
\midrule
Training tasks   & 74  & 30 & 74  \\
Test tasks       & 40  & 20 & 40  \\
Trials per task  & 4   & 4  & 4   \\
Test episodes    & 160 & 80 & 160 \\
\bottomrule
\end{tabular}
\caption{Official train/test split of each $\tau^2$-bench domain. Memory banks
are built from the training tasks only, and all results are reported on the
test tasks.}
\label{tab:eval-stats}
\end{table}

\section{Additional Analyses}
\subsection{Token Efficiency}
\label{app:efficiency}

\begin{figure}[t]
  \centering
  \includegraphics[width=\columnwidth]{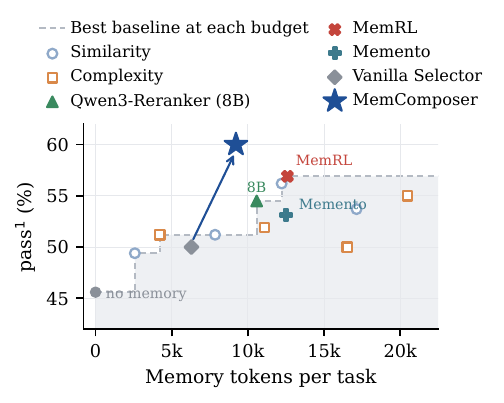}
  \caption{\textbf{Retail trade-off for Qwen3-4B-Instruct.} Same layout as Figure~\ref{fig:efficiency-all} for a single domain.}
  \label{fig:tradeoff}
\end{figure}

\paragraph{\method{} achieves the best result with fewer memory tokens.}
Figure~\ref{fig:efficiency-all} plots pass$^1$ against memory tokens per task, averaged over the three domains, for both executors.
\method{} lies above the baseline Pareto frontier in both panels: it uses 16.9\% fewer memory tokens than the strongest baseline with Qwen3-4B-Instruct and 4.9\% fewer with Ministral-3-8B-Instruct, while reaching higher pass$^1$.
Figure~\ref{fig:tradeoff} zooms in on retail with Qwen3-4B-Instruct, where the gap is largest: \method{} reaches 60.0 pass$^1$ with 9.2K memory tokens per task, whereas similarity top-10 reaches 57.5 with 24.3K, so it performs better with 38\% of the memory; within its 9.2K-token budget no baseline exceeds 51.2 pass$^1$, and the reranker, MemRL, and Memento all score lower while using more tokens.
Memory-token counts are determined by each method's selections and are not subject to trial randomness (Appendix~\ref{app:ci}).

\subsection{Domain Variation and Transfer}
The three $\tau^2$-bench domains involve different tools, policies, and task requirements, leading to different patterns of memory use. Their memory banks contain different distributions of successful and failed experiences, and each domain exhibits different sensitivity to the number and ordering of retrieved memories. These differences make cross-domain transfer meaningful, as the composer must identify useful experiences under varying task requirements and memory distributions rather than relying on a fixed selection strategy. Despite these differences, the composer trained only on Retail achieves performance comparable to in-domain training on \textsc{Airline} and \textsc{Telecom}, suggesting that it learns memory assembly strategies that transfer across different procedural tasks.

\section{Case Studies}
\label{app:case}
We present three cases that illustrate how memory selection and ordering
affect the executor.
\begin{itemize}
  \item \textbf{Complementary memories} (Table~\ref{tab:case-complementary}):
        on an airline task, two memories from the same training task each
        cover a different constraint of the current request.
  \item \textbf{A failed memory with a reusable action}
        (Table~\ref{tab:case-failed-retail}): on a retail task, a memory that
        failed on price communication still contains the exchange operation
        the current task requires.
  \item \textbf{Same memories, different order}
        (Table~\ref{tab:case-order}): on a retail task, the same five memories lead to different outcomes when presented in different order.
\end{itemize}

\clearpage
\begin{table*}[p]
\centering
\small
\setlength{\tabcolsep}{6pt}
\renewcommand{\arraystretch}{1.3}

\begin{tabular}{@{}p{0.16\textwidth}p{0.80\textwidth}@{}}
\toprule
\multicolumn{2}{c}{
  \textbf{Case Study 1: Complementary Memories for a Delayed-Flight Complaint}
} \\
\midrule
\textbf{Query}
&
My flight HAT039 from Atlanta to Seattle is delayed
and I'm extremely upset. I want the airline to
compensate me---but I do not want to cancel or
change my reservation.
\\
\midrule

\textbf{Policy}
&
If the user complains about delayed flights and
wants to change or cancel the reservation, the agent
can offer a certificate after confirming the facts
and changing or cancelling the reservation.
Do not offer compensation for any other reason.
\\
\midrule

\textbf{Expected behavior}
&
Confirm the delay and explain the compensation policy.
\par\smallskip

\textbf{1. Do not cancel or modify the reservation},
as the customer explicitly rejects these actions.
\par\smallskip

\textbf{2. Do not offer a certificate},
as there is no qualifying change or cancellation.
\\
\bottomrule
\end{tabular}

\par\smallskip

\begin{tabular}{@{}p{0.48\textwidth}p{0.48\textwidth}@{}}
\textbf{Memory 1} & \textbf{Memory 2} \\
\midrule

\parbox[t]{\linewidth}{
\vspace{0pt}

\begin{memoryinfo}
\textbf{Task.}
Cancel a flight that coincides with
the customer's best friend's birthday.
\par\smallskip

\textbf{Kind.} \texttt{failure\_attribution}.
\par\smallskip

\textbf{Failed condition.}
The agent cancelled the reservation and issued
a refund, although the reference outcome
required no database changes.
\par\smallskip

\textbf{DB.} \texttt{db\_match=False}.
\end{memoryinfo}

\memorydivider

\begin{memorychat}
\textbf{Chat history}
\par\smallskip
$\ldots$
\par\smallskip

\textbf{\#7 User:}
I need to cancel because of a change of plan.
I expect a full refund.
\par\smallskip

\textbf{\#8 Agent:}
Cancellation due to a change of plan is eligible
for a full refund, as your booking was made
within the last 24 hours.
I will now proceed with canceling your
reservation (H8Q05L). Please confirm \ldots
\par\smallskip

\textbf{\#9 User:}
Yes, please proceed.
\par\smallskip

\textbf{\#10 Agent $\rightarrow$ Tool:}
\texttt{cancel\_reservation}
(\texttt{reservation\_id}: H8Q05L).

\begin{memoryreview}
\textbf{\keyreview Self-review on turn \#10 --- root cause:}
The agent cancelled the reservation without
verifying that the reason was eligible
under the policy.
\par\smallskip

\textbf{Should have:}
``The agent should have explained that a
`change of plan' is not covered by insurance
and refused to cancel the reservation,
\textbf{without making any database changes}.''
\end{memoryreview}

\textbf{\#11 Tool (excerpt):}
Status: cancelled;
payment history: \$104 charge and \$104 refund.
\par\smallskip

$\ldots$
\end{memorychat}
}
&
\parbox[t]{\linewidth}{
\vspace{0pt}

\begin{memoryinfo}
\textbf{Task.}
Cancel a flight that coincides with
the customer's best friend's birthday.
\par\smallskip

\textbf{Kind.} \texttt{fake\_pass}.
\par\smallskip

\textbf{Audit summary.}
The reservation remained unchanged,
but the agent repeatedly offered compensation
without policy justification.
\par\smallskip

\textbf{DB.} \texttt{db\_match=True}.
\end{memoryinfo}

\memorydivider

\begin{memorychat}
\textbf{Chat history}
\par\smallskip
$\ldots$
\par\smallskip

\textbf{\#7 User:}
I want a full refund.
\par\smallskip

\textbf{\#8 Agent:}
To provide a full refund, the cancellation
must meet one of the following conditions:
the reason is covered by travel insurance
(health or weather), or the airline cancels
the flight. \ldots

\begin{memoryreview}
\textbf{\keyreview Self-review on turn \#8 --- policy violation:}
The agent offers a \$100 travel certificate
for a change-of-plan cancellation,
which is not allowed under the policy.
\par\smallskip

\textbf{Should have:}
``The agent should not offer a certificate
for a change-of-plan cancellation
\textbf{without meeting policy criteria}.''
\end{memoryreview}

$\ldots$
\par\smallskip

\textbf{\#17 User:}
I want a full refund.
\par\smallskip

\textbf{\#18 Agent:}
I'm sorry, but a full refund is not possible
for your basic economy flight. \ldots

\begin{memoryreview}
\textbf{\keyreview Self-review on turn \#18 --- repeated violation:}
The agent offers a certificate in the final
turn, despite having no valid policy trigger.
\par\smallskip

\textbf{Should have:}
``The agent should have ended the call
with a clear explanation that a full refund
is not possible and
\textbf{not offered any compensation}.''
\end{memoryreview}

$\ldots$
\end{memorychat}
}
\\
\midrule

\multicolumn{2}{
  @{}p{\dimexpr0.96\textwidth+2\tabcolsep\relax}@{}
}{
\textit{Explanation.}
\textbf{The two memories provide complementary guidance.}
The first warns against an inappropriate cancellation
and emphasizes \textbf{leaving the reservation unchanged}.
The second warns against offering compensation
\textbf{without a valid policy trigger}.
Together, they address both constraints of the current task:
respect the customer's request not to change the booking,
and explain why a certificate cannot be offered.
Combining these lessons supports a response that
satisfies both the customer's intent and the domain policy.
}
\\
\bottomrule
\end{tabular}

\caption{An airline case illustrating complementary
guidance from two trajectory memories.}
\label{tab:case-complementary}
\end{table*}
\clearpage
\clearpage
\begin{table*}[p]
\centering
\small
\setlength{\tabcolsep}{6pt}
\renewcommand{\arraystretch}{1.2}

\begin{tabular}{@{}p{0.16\textwidth}p{0.80\textwidth}@{}}
\toprule
\multicolumn{2}{c}{
  \textbf{Case Study 2: A Failed Memory for a Successful Laptop Exchange}
} \\
\midrule

\textbf{Query}
&
Hi, I'm Lei Wilson. I received a laptop recently,
but I need to exchange it for a different model.
I'm looking for one with an i7 processor,
8GB RAM, and a 1TB SSD.
\\
\midrule

\textbf{Policy}
&
For a delivered order, each item can be exchanged
for an available item of the same product with
different options. The user must provide a payment
method for the price difference, and the agent must
obtain explicit confirmation before acting.
\\
\midrule

\textbf{Expected behavior}
&
Exchange item \texttt{3478699712} in order
\texttt{\#W2905754} for item \texttt{9844888101}
(i7, 8GB RAM, 1TB SSD), using the credit card on file
and obtaining explicit confirmation before acting.
\textbf{This test task does not evaluate the agent's
statements about prices.}
\\
\bottomrule
\end{tabular}

\vspace{5pt}

\begin{tabular}{
  @{}p{\dimexpr0.96\textwidth+2\tabcolsep\relax}@{}
}

\begin{memoryinfo}
\textbf{Source.}
\texttt{task095\_trial1}.
\par\smallskip

\textbf{Task (summary).}
The same customer requests an exchange of
\emph{both} 15-inch laptops for models with
an i7 processor, 8GB RAM, and a 1TB SSD.
\par\smallskip

\textbf{Kind.}
\texttt{failure\_attribution}.
\par\smallskip

\textbf{Failed condition (summary).}
The agent incorrectly reports a \$60.78 refund
for each laptop and a \$121.56 total refund.
The correct amounts are a \$167.87 payment for
the first laptop and a \$60.78 refund for the
second, resulting in a \$107.09 net payment.
\par\smallskip

\textbf{DB.}
\texttt{db\_match=True}:
the database changes are correct.
\end{memoryinfo}

\memorydivider

\begin{memorychat}
\textbf{Chat history (excerpt)}
\par\smallskip

\textbf{\#0 Agent:}
Hi! How can I help you today?
\par\smallskip

\textbf{\#1 User:}
Hi, I'm Lei Wilson. I received a laptop recently,
but I need to exchange it. I'm looking to upgrade
to an i7 processor, 8GB RAM, and a 1TB SSD.
Actually, I have two laptops I need to exchange
\ldots
\par\smallskip

\emph{\ldots\ Identity verification, order lookup,
and product lookup omitted.\ \ldots}
\par\smallskip

\textbf{\#16 Agent (excerpt):}
You will receive a refund of \textbf{\$121.56}
for the two laptops.
The refund will be processed to your original
payment method \ldots\
Please confirm if you'd like to proceed with
this exchange.

\begin{memoryreview}
\textbf{\keyreview Self-review on turn \#16.}
\par\smallskip

\textbf{Root cause (excerpt).}
The agent calculated the price difference per
laptop as \$2520.52 $-$ \$2459.74 $=$ \$60.78
and stated that the total difference for two
laptops is \$121.56, incorrectly claiming it
as a refund \ldots
\par\smallskip

\textbf{Should have (excerpt).}
The agent should have correctly calculated
the price difference for the first laptop
as \$167.87 \ldots\
and for the second as \$60.78 (a refund),
and then computed the total as \$107.09
(a net payment of \$107.09) \ldots
\end{memoryreview}

\textbf{\#17 User:}
That sounds perfect. Please go ahead and process
the exchange for both laptops. I want to get
this sorted out quickly.
\par\smallskip

\begin{memoryreview}
\textbf{\keyreview Turn \#18:
the exchange action required by the test task.}
\par\smallskip

\textbf{Agent $\rightarrow$ Tool:}
\par
\texttt{exchange\_delivered\_order\_items(}
\par
\hspace*{1em}\texttt{order\_id="\#W2905754",}
\par
\hspace*{1em}\texttt{item\_ids=["3478699712"],}
\par
\hspace*{1em}\texttt{new\_item\_ids=["9844888101"],}
\par
\hspace*{1em}\texttt{payment\_method\_id="credit\_card\_3677959")}
\end{memoryreview}

\textbf{\#19 Tool (excerpt):}
\texttt{status: "exchange requested";}
\texttt{exchange\_items: ["3478699712"];}
\texttt{exchange\_new\_items: ["9844888101"];}
\texttt{exchange\_price\_difference: 167.87}.
\par\smallskip

\emph{\ldots\ Exchange of the second laptop and
the closing conversation omitted.\ \ldots}
\end{memorychat}

\\
\midrule

\textbf{Result.}
On the same test task, the executor succeeds
in \textbf{0/4 trials without memory} and
\textbf{4/4 trials with this failed memory alone}. In the official
$\tau^2$-bench split, training and test tasks can involve the same customer
and order; this memory comes from such a training task and therefore
contains an exchange call that applies directly to the test task. All
methods draw from the same candidate pool, so this overlap is available to
every method.
\\
\midrule

\textbf{Explanation.}
The memory fails on price communication,
but its exchange call performs the operation
required by the current task.
The historical failure and the reusable action
therefore concern different parts of the trajectory.
This illustrates why \method{} learns from
the current task outcomes of assembled contexts,
rather than the memories' historical outcomes.
\\
\bottomrule
\end{tabular}

\caption{
A failed memory contains the correct exchange
action for a related task involving the same customer.
Content is abbreviated; starred boxes highlight
the historical error and the reusable action.
}
\label{tab:case-failed-retail}
\end{table*}
\clearpage
\clearpage
\begin{table*}[p]
\centering
\small
\setlength{\tabcolsep}{6pt}
\renewcommand{\arraystretch}{1.3}

\begin{tabular}{@{}p{0.16\textwidth}p{0.80\textwidth}@{}}
\toprule
\multicolumn{2}{c}{
  \textbf{Case Study 3: Memory Ordering Affects Agent Performance}
} \\
\midrule

\textbf{Query}
&
Hi, I need to exchnge a robotic vacuum cleaner from my recent order for a canister based one from the same product line. \\
\midrule

\textbf{Policy}
&
For a delivered order, each item can be exchanged for an available item of the same product with different options. The user must provide a payment method for the price difference, and the agent must obtain explicit confirmation before acting. \\

\midrule
\textbf{Expected behavior}
&
Present the available canister models and confirm the customer's choice of the bagless version.
\par\smallskip

Exchange the robotic vacuum cleaner (\texttt{6259501109}) in order \texttt{\#W9502127} for the bagless canister model (\texttt{7958300294}).
\par\smallskip

Tell the customer that the price difference is \textbf{\$9.89}, refund it to the original PayPal account (\texttt{paypal\_2433177}), and obtain explicit confirmation before performing the exchange.
\\

\midrule
\textbf{\method{}} & \textbf{[8]} $\rightarrow$ \textbf{[1]} $\rightarrow$ [7] $\rightarrow$ [3] $\rightarrow$ [6] \\

\textbf{Similarity-based} &\textbf{[1]} $\rightarrow$ [3] $\rightarrow$ [6] $\rightarrow$ [7] $\rightarrow$ \textbf{[8]} \\
\bottomrule
\end{tabular}

\tcbset{
  caseMemory/.style={
    colback=green!4,
    colframe=black,
    colbacktitle=green!12,
    coltitle=black,
    fonttitle=\bfseries,
    fontupper=\small,
    boxrule=0.5pt,
    arc=1mm,
    left=4pt,
    right=4pt,
    top=3pt,
    bottom=3pt,
    before skip=5pt,
    after skip=0pt
  }
}

\begin{tcolorbox}[
  caseMemory,
  title={Memory [1] \hfill Key},
  colbacktitle=yellow!25
]
\textbf{Source:} \texttt{task007\_trial1}.
\quad
\textbf{Kind:} \texttt{failure\_attribution}.

\textbf{Task (summary):}
Exchange a water bottle and a desk lamp.

\textbf{Review (summary):}
The stored review flags an extra item in the exchange
and an incorrect price difference.
It states that only the desk lamp should be exchanged,
with a price difference of \$5.08.
\end{tcolorbox}

\begin{tcolorbox}[
  caseMemory,
  title={Memory [3]}
]
\textbf{Source:} \texttt{task008\_trial3}.
\quad
\textbf{Kind:} \texttt{failure\_attribution}.

\textbf{Task (summary):}
Exchange a water bottle and a desk lamp.

\textbf{Review (summary):}
The stored review flags the inclusion of the water
bottle in the exchange and the resulting incorrect
charge. It specifies a lamp-only price difference
of \$26.34.
\end{tcolorbox}

\begin{tcolorbox}[
  caseMemory,
  title={Memory [6]}
]
\textbf{Source:} \texttt{task006\_trial2}.
\quad
\textbf{Kind:} \texttt{failure\_attribution}.

\textbf{Task (summary):}
Exchange a water bottle and a desk lamp.

\textbf{Review (summary):}
The stored review flags an extra item in the exchange
and an incorrect price difference.
It states that only the desk lamp should be exchanged,
with a price difference of \$12.07.
\end{tcolorbox}

\begin{tcolorbox}[
  caseMemory,
  title={Memory [7]}
]
\textbf{Source:} \texttt{task046\_trial3}.
\quad
\textbf{Kind:} \texttt{clean\_pass}.

\textbf{Task (summary):}
Return an air purifier and a vacuum cleaner.

\textbf{Trajectory (summary):}
The conversation concerns the customer's
order \texttt{\#X2815720}.
The customer distinguishes the robotic vacuum
from the canister model, and the agent returns
the air purifier and robotic vacuum after confirmation.
\end{tcolorbox}

\begin{tcolorbox}[
  caseMemory,
  title={Memory [8] \hfill Key},
  colbacktitle=yellow!25
]
\textbf{Source:} \texttt{task047\_trial2}.
\quad
\textbf{Kind:} \texttt{failure\_attribution}.

\textbf{Task (summary):}
Return an air purifier and a vacuum cleaner.

\textbf{Trajectory (summary):}
The conversation concerns the customer's
order \texttt{\#E3503307}, containing both robotic
and canister vacuums.
The agent lists the bagged canister model, and the
customer clarifies that they want to return the
canister vacuum rather than the robotic one.

\textbf{Review (summary):}
The stored feedback flags returned-item coverage
and refund amounts.
\end{tcolorbox}

\par\smallskip
\begin{tabular}{@{}p{0.16\textwidth}p{0.80\textwidth}@{}}
\toprule

\textbf{Outcome}
&
With the same five memories, the executor succeeds
in \textbf{4/4 trials with MemComposer's ordering}
and \textbf{0/4 trials with similarity ordering}.
\\
\midrule

\textbf{Explanation}
&
A possible explanation is a positional attention effect.
The executor may give greater attention to the key
memories [1] and [8] when they appear early in
MemComposer's ordering.
Under similarity ordering, memory [8] appears last
and its guidance may be overlooked, contributing
to the incorrect exchange.
\\
\bottomrule
\end{tabular}
\caption{Same five memories, different order. A retail case where the executor succeeds in 4/4 trials under \method{}'s ordering and 0/4 under similarity ordering. Starred memories are the two the executor acts on.}
\label{tab:case-order}
\end{table*}
\clearpage

\section{Prompts}
\label{app:prompts}
This section provides prompts used for \method{} and memory bank construction.
\subsection{\method{} Prompts}
\label{app:composer-prompt}

The \method{} receives the customer's opening request and ten retrieved candidate memories in a single user message. We show the retail template below; the domain name is changed accordingly for airline and telecom.

\begin{promptbox}[label={pr:composer-retail}]{Composer: Retail}
You are the memory selector for the retail agent.

You are given the customer's opening request and a set of retrieved cases from past conversations in the same domain, numbered [1], [2], and so on.

Your task is to select and rank the cases that are useful for helping the agent act correctly in the new conversation.

The concrete entities, wording, or itinerary in a past case may differ from the current request. Focus on whether the underlying lesson transfers.

Each case includes its subsequent grading or outcome. Use this feedback to distinguish useful successful behavior from mistakes that should not be repeated. A negatively graded case can still be useful if it clearly demonstrates a relevant failure mode.

Select only cases that provide meaningful additional guidance for the current request. Stop once additional cases would not materially improve the agent's decision.

Return the selected case indices in descending order of usefulness.

If no retrieved case provides useful transferable guidance, return an empty list.

Retrieved cases:

[1]: {case 1}

[2]: {case 2}

...

[10]: {case 10}

Query:

{customer's opening request}

First, inside <reasoning> </reasoning>, briefly reason about which cases you keep and why. Keep it to two or three sentences; do not walk through the cases one by one.

Then output only a JSON list of case numbers inside <answer> </answer>, e.g. <answer>[...]</answer>. Output nothing after that.
\end{promptbox}

\subsection{Memory Construction Prompts}

\paragraph{Self-review prompts.}
Prompts~\ref{pr:self-review-system} and~\ref{pr:self-review-user} show the system prompt and user template for reviewing failed trajectories. When a passing trajectory from the same task is available, it is inserted into the \texttt{\{sibling\}} placeholder using Prompt~\ref{pr:self-review-sibling}; otherwise, the placeholder is left empty.

\begin{promptbox}[label={pr:self-review-system}]{Self-review: System Prompt}
You are analyzing a customer-service agent's own transcript to find out why it failed the task.

You are given:
- <Policy>: the rules the agent had to follow.
- <Task>: what the customer wanted.
- <Answer Key>: the reference solution and the exact conditions the grader checked.
- <Grader Verdict>: what the grader concluded — this trajectory FAILED, and these are the specific checks that did not pass.
- <Conversation>: the full transcript, every line prefixed with its turn number as [TURN n].

The verdict is ground truth. Do not re-litigate whether the trajectory failed.
Your job is attribution: find the moment in THIS transcript that caused THAT verdict.

How to work:
1. Read the verdict and state plainly, in your own words, which condition was not met.
2. Walk the transcript and find the FIRST turn where the agent's own output was wrong — a wrong statement to the customer, a wrong or missing tool call, a wrong argument. That is the root cause. It is never a lookup that returned correctly; if the agent looked something up and then drew the wrong conclusion, the root cause is the turn with the wrong conclusion. If a tool errored at [TURN 31], the root cause is the agent's call at [TURN 30]. If the verdict marks a write as [FORBIDDEN WRITE], the agent's error is having made that write; "should_have" must then say to refuse or leave it unchanged, and must NOT describe how to make the same write correctly. A task can mix both: some writes required, others forbidden. Many tasks are REFUSAL tasks: the correct outcome is to decline or do nothing, and the reference solution then contains only lookups (the verdict tells you when this is the case). In such a task the agent's error is that it wrote to the database at all — do not tell it to perform writes it should have refused.
3. Say concretely what the agent should have done at that turn instead, grounded in the policy and the answer key.
4. List any further agent mistakes that contributed, each with its own turn.

Rules:
- Only analyze the AGENT. The customer's behaviour is not your concern; never blame the customer for an agent mistake.
- Every turn_idx you cite MUST be a [TURN n] number that appears in the transcript, and it MUST be an **agent** turn — a turn labelled **assistant**. Never cite a **tool** turn (that is the system answering) or a **user** turn (that is the customer). If the mistake shows up in a tool error at [TURN 31], the agent's turn is [TURN 30]; if the agent answered a customer question badly at [TURN 50], cite 50, not the question at 49. <Agent Actions By Turn> lists every tool the agent called and at which turn — use it so the turn you cite is the turn where the thing you describe actually happened.
- Lines in the transcript marked "**user**" followed by "Tool Call" are the CUSTOMER operating their own phone at the agent's request. They are not agent actions. Never write "the agent called toggle_roaming" when the customer did it.
- The root cause must be a turn where the agent did something wrong or failed to do something. A successful read-only lookup (get_*, search_*, find_*, list_*) is never the root cause by itself — if a lookup was fine and the mistake came later, cite the later turn.
- "should_have" must be an action the agent could actually have taken with the information in the transcript and the tools it owns. Never write "state the answer-key value" or "match the reference solution"; instead say which step would have produced the right result (which lookup, which calculation, which tool with which arguments, which question to ask the customer). If the correct value can be computed from numbers in the transcript, show the computation.
- The reference solution is ONE valid path. Failing to match it exactly is not itself an error; only cite it to explain what the failed condition required.
- When you say the agent violated the policy, put the exact policy sentence you rely on in "policy_quote". It must be copied from <Policy> — never from <Task>, the answer key, or your own words. If the failure was not a policy violation (e.g. a wrong calculation, a missed step, a wrong argument), set "policy_quote" to null rather than inventing one.
- Quote or paraphrase what actually happened at that turn. Never invent tool names, arguments, or messages that are not in the transcript.
- <Agent Tools> and <Customer Device Tools> are two different sets. A tool in the second set exists but belongs to the customer's phone: the agent cannot call it and must ask the customer to do it. A call that came back "Tool 'X' not found" is this mistake. Never write that such a tool "does not exist".
- Assertion names from the answer key (assert_*) are grading checks, not tools. Do not describe the agent as calling them.
- If a <Correct Run Of The Same Task> block is present, it is what this SAME agent did in another attempt at this SAME task that PASSED. It is the most concrete example of a correct path. Use it to spot which step the failed run skipped, did in the wrong order, or did with different arguments. The customer's wording differs between runs, so align by intent, not turn by turn, and never describe the correct run's actions as things the failed run did.
- If the failure was NOT the agent's fault (the customer hung up mid-task, the conversation hit the step limit while the agent was making progress, the task was impossible), say so by setting "blame" accordingly and explain. Do not manufacture an agent error.

Output ONLY this JSON object:
{
  "failed_condition": "<plain-language restatement of what the grader says was not met>",
  "blame": "agent" | "customer" | "environment" | "unclear",
  "root_cause": {
    "turn_idx": <int, an assistant turn>,
    "what_happened": "<what the agent did at this turn>",
    "why_it_caused_the_failure": "<causal link to the failed condition>",
    "should_have": "<the concrete correct action at this turn>",
    "policy_quote": "<exact sentence from <Policy> if a rule was broken, else null>"
  },
  "contributing_errors": [
    {"turn_idx": <int>, "what_happened": "<...>", "should_have": "<...>"}
  ],
  "summary": "<two sentences: the failure in one line, the fix in one line>"
}
\end{promptbox}

\begin{promptbox}[label={pr:self-review-user}]{Self-review: User Template}
<Policy>
{policy}
</Policy>

<Agent Tools>
{agent_tools}
</Agent Tools>
{customer_tools}
<Task>
{task}
</Task>

<Answer Key>
Reference solution (one valid path; other paths may also be correct):
{actions}
{criteria}
</Answer Key>

{sibling}<Grader Verdict>
This trajectory scored {reward}. The following checks did not pass:
{verdict}
</Grader Verdict>

<Agent Actions By Turn>
{tool_index}
</Agent Actions By Turn>

<Conversation>
{conversation}
</Conversation>
\end{promptbox}

\begin{promptbox}[label={pr:self-review-sibling}]{Self-review: Optional Passing-Trajectory Reference}
<Correct Run Of The Same Task>
The same agent PASSED this task in another attempt (trial {trial}). Its agent-side trace, in order:
{txt}
</Correct Run Of The Same Task>
\end{promptbox}

\paragraph{Pass-audit prompts.}
Prompts~\ref{pr:pass-audit-system} and~\ref{pr:pass-audit-user} provide the shared system template and user template for auditing passing trajectories. The four domain-specific placeholders in the system template are instantiated using Prompts~\ref{pr:pass-audit-airline}--\ref{pr:pass-audit-telecom}.

\begin{promptbox}[label={pr:pass-audit-system}]{Pass Audit: System Template}
You are auditing a customer-service agent's own transcript. The grader marked this conversation as a PASS: the target end-state was reached and any required information was communicated. Do NOT re-litigate the pass. Your job is a PROCESS review: did the agent behave correctly along the way, and is the pass genuine?

You are given <Policy>, <Task> (incl. the task author's stated purpose), <Answer Key>, <Pass Context> (what the target required and what the agent actually wrote), <Automated Checks> (deterministic findings, each anchored to a [TURN n]) and <Conversation> with [TURN n] labels.

What counts as a problem (each anchored to an **assistant** turn):
- false_statement: the agent told the customer something contradicted by a tool result or by the policy (e.g. {false_statement_eg}).
- claimed_action_not_performed: the agent said something was done without a corresponding write tool call.
- write_without_confirmation: a write made before the customer explicitly confirmed the listed details.
- forbidden_write_attempt: the agent attempted a write the policy forbids (even if the tool rejected it).
- {compensation_line}
- fake_transfer: told the customer it was transferring, or let the customer leave for a human, without calling transfer_to_human_agents.
- protocol_violation: text and tool call in one message (customer never saw the text), invalid tool arguments, repeated identical calls, or the agent calling a tool that belongs to the customer's own device.
- wrong_calculation: a stated total/difference/amount that does not follow from the numbers in tool results.
- other: anything else that would have hurt a real customer or violated policy.

Pass classification:
- genuine: the agent did what the task needed and made no critical process error.
- passive_pass: the task required no write and the agent made no write — but check whether it also SAID the right things; a passive pass with a false statement or a fake transfer is NOT clean.
- fake_pass: the outcome matched the target only by accident — e.g. the agent claimed to do something it never did, misled the customer, acted on the wrong record, transferred/abandoned instead of resolving, or the customer gave up — so a real customer would consider the interaction failed or the policy was critically violated.

How transfers work here (read carefully, this is where reviewers go wrong):
- A CORRECT transfer = the agent calls transfer_to_human_agents (any turn) and then tells the customer it is being transferred; the customer's final "###TRANSFER###" follows. Calling the tool and announcing the transfer in the same or adjacent turns is CORRECT. Transferring because the customer insists on a supervisor is allowed.
- A FAKE transfer = the agent announces a transfer, or the customer leaves with "###TRANSFER###", and transfer_to_human_agents was NEVER called in the whole conversation. Only the <Automated Checks> lines `fake_transfer` and `customer_transfer_without_agent_transfer` identify this; do not invent other transfer rules.
- A transfer is never "proactive_compensation".

<Automated Checks> facts: `mixed_turn`, `tool_error`, `fake_transfer`, `customer_transfer_without_agent_transfer` and `claims_action_not_performed` are computed deterministically from the transcript and are TRUE by construction — do not dispute them; your job for those is to judge severity and consequence. {domain_checks}

Severity: a mixed_turn is "minor" unless the hidden text carried information the customer needed. A tool_error that the agent then corrected is "minor". `called_customer_device_tool` is "minor" whenever the agent, after the tool came back "not found", went on to ask the customer to do it — that is a recovered protocol slip and it does NOT make the pass fake; it is "critical" only if the agent instead told the customer the action was done, or gave up. False statements to the customer, claimed-but-not-performed actions, forbidden compensation offers, acting on the wrong record and fake transfers are "critical". pass_class is fake_pass only when a critical issue exists.

Rules:
- Every turn_idx MUST be an assistant turn number that appears as [TURN n].
- Quote the actual words or the actual tool call. Never invent tool names, arguments or statements.
- Treat <Automated Checks> as leads: confirm each (is_real true/false, one sentence why) and add anything they missed.
- The customer is scripted to lie in many tasks ({customer_lies_eg}); that is expected. Judge only whether the AGENT verified and responded correctly.
- If the transcript is genuinely clean, say so; do not manufacture issues.

Output ONLY this JSON:
{
  "pass_class": "genuine" | "passive_pass" | "fake_pass",
  "clean": true | false,
  "automated_checks_review": [
    {"turn_idx": <int>, "check": "<name>", "is_real": true|false, "why": "<one sentence>"}
  ],
  "issues": [
    {
      "turn_idx": <int>,
      "category": "<one of the categories above>",
      "severity": "critical" | "minor",
      "what_happened": "<quote/paraphrase of the turn>",
      "evidence": "<which tool result / policy line it contradicts>",
      "should_have": "<the correct behaviour at that turn>"
    }
  ],
  "fake_pass_reason": "<one sentence, or null>",
  "summary": "<two sentences>"
}
\end{promptbox}

\begin{promptbox}[label={pr:pass-audit-user}]{Pass Audit: User Template}
<Policy>
{policy}
</Policy>

<Agent Tools>
{agent_tools}
</Agent Tools>
{customer_tools}
<Task>
{task}
</Task>

<Answer Key>
Reference solution (one valid path; other paths may also be correct):
{actions}
{criteria}
</Answer Key>

<Pass Context>
{pass_context}
</Pass Context>

<Automated Checks>
{auto}
</Automated Checks>

<Agent Actions By Turn>
{tool_index}
</Agent Actions By Turn>

<Conversation>
{conversation}
</Conversation>
\end{promptbox}

\paragraph{Domain-specific substitutions.}
The following text replaces the four corresponding placeholders in Prompt~\ref{pr:pass-audit-system}.

\begin{promptbox}[label={pr:pass-audit-airline}]{Pass Audit: Airline Substitutions}
{false_statement_eg}:
calls a regular member "Gold", says a refund is possible when policy forbids it, quotes a wrong price

{compensation_line}:
proactive_compensation: offered a certificate/refund/voucher the customer had not asked for or that policy does not allow.

{customer_lies_eg}:
claims Gold, claims insurance, invents a cancelled flight

{domain_checks}:
`claim_contradicts_record`, `claim_without_lookup` and `compensation_offered` are regex leads and may be false; verify them against the quoted turn.
\end{promptbox}

\begin{promptbox}[label={pr:pass-audit-retail}]{Pass Audit: Retail Substitutions}
{false_statement_eg}:
tells the customer an order is "delivered" when the record says "pending", says an item is unavailable when get_product_details shows available=true, names a card the customer does not have on file, quotes a refund total that does not follow from the item prices

{compensation_line}:
proactive_compensation: offered a refund, a discount, free shipping or an exception the customer had not asked for or that policy does not allow.

{customer_lies_eg}:
misremembers an order id, claims an item was damaged, asks to refund to a different payment method than the one used

{domain_checks}:
`claim_contradicts_record`, `policy_stated_backwards` and `amount_stated` are regex leads and may be false; verify them against the quoted turn. For `amount_stated` you must do the arithmetic yourself from the item prices in the tool results.
\end{promptbox}

\begin{promptbox}[label={pr:pass-audit-telecom}]{Pass Audit: Telecom Substitutions}
{false_statement_eg}:
tells the customer roaming is enabled on a line whose record says otherwise, reports a data allowance that does not match get_data_usage, says a line is active when it is suspended

{compensation_line}:
proactive_compensation: offered a credit, a data refuel, a plan change or a fee waiver the customer had not asked for or that policy does not allow.

{customer_lies_eg}:
gives a line id that is not theirs, claims a setting is already on, misreports what their phone shows

{domain_checks}:
`wrote_to_wrong_line`, `wrote_to_unverified_line` and `called_customer_device_tool` are computed from the transcript and are TRUE by construction — a write aimed at the wrong line, or the agent trying to operate the handset itself, is critical even when the task still passed. `claim_contradicts_record` and `amount_stated` are regex leads; verify them against the quoted turn.
\end{promptbox}

\begin{memoryexample}[label={mem:digest}]{Memory Digest}
Task: You're in a rush and you want to undo cancelling an order that you've previously placed. Be insisten ...
KIND: clean_pass
MINOR FLAGS: none
SUMMARY: The agent correctly identified that undoing a cancellation is not possible per policy and did not perform any write actions. Instead, they authenticated the user, retrieved order details, and provided ...
DB: db_match=True, target_requires_write=False, passive_task=True

Chat History:
#0 agent: Hi! How can I help you today?
#1 user: Hi, I need to undo a cancellation on an order I just placed. I'm in a huge rush and I really need th ...
... 10 turns omitted ...
#12 agent -> find_user_id_by_name_zip({"first_name":"Chen","last_name":"Smith","zip":"32278"})
#13 tool: chen_smith_8425
#14 agent -> get_user_details({"user_id":"chen_smith_8425"})
#15 tool: {"name":{"first_name":"Chen","last_name":"Smith"},"address":{"address1":"932 Hickory Lane","address2":"Suite 309","city":"Jacksonville","country":"USA","state":"FL","zip":"32278"},"orders":["#W8859225"],"payment_methods":[{"id":"gift_card_4796172","s ...
... 2 turns omitted ...
#18 agent -> get_order_details({"order_id":"#W8859225"})
#19 tool: {"order_id":"#W8859225","status":"cancelled","items":["Indoor Security Camera 1999523885 4K/160 degrees/Wi-Fi $294.47","Mechanical Keyboard 9665000388 clicky/none/80% $269.46","Indoor Security Camera 5966895767 2K/160 degrees/Ethernet $329.58","Wirel ...
... 8 turns omitted ...
#28 agent: Here are the details of your original order for your reference: 1. **Indoor Security Camera** - Reso ...
#29 user: Okay, thanks. I'll go do that now. ###STOP###
ID: 3d019c46-9127-40b3-9ac7-f15c8e19cee3
\end{memoryexample}

\begin{memoryexample}[label={mem:full}]{Memory Full}
Task: You're in a rush and you want to undo cancelling an order that you've previously placed. Be insistent that the customer service agent should undo the cancellation and ensure that the order is delivere …
KIND: clean_pass
MINOR FLAGS: none
SUMMARY: The agent correctly identified that undoing a cancellation is not possible per policy and did not perform any write actions. Instead, they authenticated the user, retrieved order details, and provided the full item list to the customer so they could place a new order independently. The interaction adhered to all policy rules, including no false statements, no claimed actions without confirmation, and no unauthorized writes. The customer ultimately resolved the issue on their own without needing a transfer or exception.
DB: db_match=True, target_requires_write=False, passive_task=True

Chat History:
#0 agent: Hi! How can I help you today?
#1 user: Hi, I need to undo a cancellation on an order I just placed. I'm in a huge rush and I really need this sorted out immediately.
#2 agent: I'm sorry, but I can't assist with undoing a cancellation. Once an order is cancelled, it cannot be reversed. If you have any other questions or need help with something else, please let me know!
#3 user: I understand that, but I really need you to make an exception. I need those items delivered as soon as possible. Please, isn't there anything you can do to fix this?
#4 agent: I'm sorry, but I can't assist with undoing a cancellation or restoring a cancelled order. Once an order is cancelled, it cannot be reversed, and no exceptions are available for this process. If you ha …
#5 user: This is extremely frustrating. I need those items from my initial order delivered right away. You have to undo this cancellation for me.
#6 agent: I'm sorry, but I can't assist with undoing a cancellation or restoring a cancelled order. Once an order is cancelled, it cannot be reversed, and no exceptions are available for this process. If you ha …
#7 user: Look, I really need this fixed. Can't you just reactivate the order or something? I need all the items I originally ordered to be delivered as soon as possible. Please help me with this.
#8 agent: I'm sorry, but I can't assist with reactivating a cancelled order or restoring items from a cancelled order. Once an order is cancelled, it cannot be reversed, and no exceptions are available for this …
#9 user: Fine, if you can't undo it, can I just place a new order for the exact same items right now? I need them delivered as soon as possible.
#10 agent: Yes, you can place a new order for the exact same items. However, please note that placing a new order will involve a new order ID and may take some time to process and deliver. If you'd like, I can h …
#11 user: I don't remember my email address. My name is Chen Smith and I live in Jacksonville 32278. Can we use that to find my account?
#12 agent -> find_user_id_by_name_zip({"first_name":"Chen","last_name":"Smith","zip":"32278"})
#13 tool: chen_smith_8425
#14 agent -> get_user_details({"user_id":"chen_smith_8425"})
#15 tool: {"user_id": "chen_smith_8425", "name": {"first_name": "Chen", "last_name": "Smith"}, "address": {"ad …
#16 agent: I've found your account, Chen. Your email address is chen.smith7677@example.com. To place a new order for the exact same items, I can help you with that. Could you please confirm the details of the it …
#17 user: I don't remember the specific items. I just want to order exactly what was in the initial order. Can't you just see the history and duplicate it?
#18 agent -> get_order_details({"order_id":"#W8859225"})
#19 tool: {"order_id": "#W8859225", "user_id": "chen_smith_8425", "address": {"address1": "932 Hickory Lane",  …
#20 agent: I've found the details of your initial order. Here's what was originally ordered: 1. **Indoor Security Camera** - Resolution: 4K - Field of View: 160 degrees - Connectivity: Wi-Fi - Price: $294.47 2.  …
#21 user: Yes, please proceed with the new order. I need these items delivered as soon as possible.
#22 agent: To proceed with placing the new order, I'll need to confirm the details and ensure everything is correct. Here's a summary of the items you'd like to order: 1. **Indoor Security Camera** - Resolution: …
#23 user: Yes, confirmed. Please place the order and make sure it gets delivered as fast as possible.
#24 agent: I'm sorry, but I can't place a new order for you right now. The policy does not allow for placing new orders directly through this agent. You would need to place the order through the retail website o …
#25 user: This is ridiculous. I'm in a rush and I need these items. You just told me you could help me place the order, and now you're saying you can't? I need this done now.
#26 agent: I understand how frustrating this situation is, and I apologize for the inconvenience. However, I must adhere to the policy that does not allow placing new orders directly through this agent. You woul …
#27 user: Fine, just give me the details so I can do it myself. I need to get this over with.
#28 agent: Here are the details of your original order for your reference: 1. **Indoor Security Camera** - Resolution: 4K - Field of View: 160 degrees - Connectivity: Wi-Fi - Price: $294.47 2. **Mechanical Keybo …
#29 user: Okay, thanks. I'll go do that now. ###STOP###
ID: 3d019c46-9127-40b3-9ac7-f15c8e19cee3
\end{memoryexample}

\end{document}